\documentclass[runningheads]{llncs}
\usepackage{subfigure}
\usepackage{graphicx}
\usepackage{amsmath,amssymb} 
\usepackage{color}
\usepackage{epstopdf}
\usepackage{multirow}
\usepackage[width=122mm,left=12mm,paperwidth=146mm,height=193mm,top=12mm,paperheight=217mm]{geometry}
\begin{document}
\pagestyle{headings}
\mainmatter

\title{3D Human Pose Estimation Using Convolutional Neural Networks with 2D Pose Information} 

\titlerunning{3D Human Pose Estimation Using CNNs with 2D Pose Information}

\authorrunning{S. Park et al.}

\author{Sungheon Park, Jihye Hwang, and Nojun Kwak}


\institute{Graduate School of Convergence Science and Technology,\\
	Seoul National University, Seoul, Korea\\
	\email{ sungheonpark@snu.ac.kr, hjh881120@gmail.com, nojunk@snu.ac.kr}
}

\maketitle

\begin{abstract}
While there has been a success in 2D human pose estimation with convolutional neural networks (CNNs), 3D human pose estimation has not been thoroughly studied. In this paper, we tackle the 3D human pose estimation task with end-to-end learning using CNNs. Relative 3D positions between one joint and the other joints are learned via CNNs. The proposed method improves the performance of CNN with two novel ideas. First, we added 2D pose information to estimate a 3D pose from an image by concatenating 2D pose estimation result with the features from an image. Second, we have found that more accurate 3D poses are obtained by combining information on relative positions with respect to multiple joints, instead of just one root joint. Experimental results show that the proposed method achieves comparable performance to the state-of-the-art methods on Human 3.6m dataset.
\end{abstract}

\section{Introduction}
Both 2D and 3D human pose recovery from images are important tasks since the retrieved pose information can be used to other applications such as action recognition, crowd behavior analysis, markerless motion capture and so on. However, human pose estimation is a challenging task due to the dynamic variations of a human body. Various skin colors and clothes also make the estimation difficult. Especially, pose estimation from a single image requires a model that is robust to occlusion and viewpoint variations.

Recently, 2D human pose estimation achieved a great success with convolutional neural networks (CNNs)~\cite{toshev2014deeppose,carreira2015human,wei2016convolutional}. Strong representation power and the ability to disentangle underlying factors of variation are characteristics of CNNs that enable learning discriminative features automatically~\cite{bengio2009learning} and show superior performance to the methods based on hand-crafted features. On the other hands, 3D human pose estimation using CNNs has not been studied thoroughly compared to the 2D cases. Estimating a 3D human pose from a single image is more challenging than 2D cases due to the lack of depth information. However, CNN can be a powerful framework for learning discriminative image features and estimating 3D poses from them. In the case where the target object is fixed such as human body, it is able to learn useful features directly from images without keypoint matching step in the typical 3D reconstruction tasks.

Though recent algorithms that are based on CNNs for 3D human pose estimation have been proposed~\cite{li20143d,li2015maximum,tekin2016structured}, they do not make use of 2D pose information which can provide additional information for 3D pose estimation. From 2D pose information, undesirable 3D joint positions which generate unnatural human pose may be discarded. Therefore, if the information that contains the 2D position of each joint in the input image is used, the results of 3D pose estimation can be improved.

In this paper, we propose a simple yet powerful 3D human pose estimation framework based on the regression of joint positions using CNNs. We introduce two strategies to improve the regression results from the baseline CNNs. Firstly, not only the image features but also 2D joint classification results are used as input features for 3D pose estimation. This scheme successfully incorporates the correlation between 2D and 3D poses. Secondly, rather than estimating relative positions with respect to only one root joint, we estimated the relative 3D positions with respect to multiple joints. This scheme effectively reduces the error of the joints that are far from the root joint. Experimental results validate the proposed framework significantly improves the baseline method and achieves comparable performance to the state-of-the-art methods on Human 3.6m dataset~\cite{h36m_pami} without utilizing the temporal information.

The rest of the paper is organized as follows. Related works are reviewed in Section~\ref{sec:rel}. The structure of CNNs used in this paper and two key ideas of our method, 1) the integration of 2D joint classification results into 3D pose estimation and 2) multiple 3D pose regression from various root nodes, are explained in Section~\ref{sec:cnn}. Details of implementation and training procedures are explained in Section~\ref{sec:detail}. Experimental results are illustrated in Section~\ref{sec:exp}, and finally conclusions are made in Section~\ref{sec:con}.

\section{Related Work}
\label{sec:rel}
Human pose estimation has been a fundamental task since early computer vision literature, and numerous researches have been conducted on both 2D and 3D human pose estimation. In this section, we will cover both 2D and 3D human pose estimation methods focusing on the CNN-based methods.

Early works for 2D human pose estimation which are based on deformable parts model~\cite{felzenszwalb2005pictorial}, pictorial structure~\cite{andriluka2009pictorial,dantone2013human,yang2011articulated}, or poselets~\cite{bourdev2009poselets} train the relationship between body appearance and body joints using hand-crafted features. Recently proposed CNN based methods drastically improve the performance over the previous hand-crafted feature based methods. DeepPose~\cite{toshev2014deeppose} used CNN-based structure to regress joint locations with multiple iterations. Firstly, it predicts an initial pose using holistic view and refine the currently predicted pose using relevant parts of the image. Xiaochuan et al.~\cite{fan2015combining} integrated both the local part appearance and the holistic view of an image using dual-source CNN. Convolutional pose machine~\cite{wei2016convolutional} is a systematic approach to improve prediction of each stage. Each stage operates a CNN which accepts both the original image and confidence maps from preceding stages as an input. The performance is improved by combining the joint prediction results from the previous step with features from CNN. Joao et al.~\cite{carreira2015human} proposed a self-correcting method by a top-down feedback. It iteratively learns a human pose using a self-correcting CNN model which gradually improves the initial result by feeding back error predictions. Xiao et al.~\cite{chu2016structured} proposed an end-to-end learning system which captures the relationships among feature maps of joints. Geometrical transform kernels are introduced to learn features and their relationship jointly.

Similar to the 2D case, early stage of 3D human pose estimation is also based on the low-level features such as local shape context~\cite{agarwal2006recovering} or segmentation results~\cite{ionescu2011latent}. With the extracted features, 3D pose estimation is formulated as a regression problem using relevance vector machines~\cite{agarwal2006recovering}, structured SVMs~\cite{ionescu2011latent}, or random forest classifiers~\cite{shotton2013real}. Recently, CNNs have drew a lot of attentions also for the 3D human pose estimation tasks. Since search space in 3D is much larger than 2D image space, 3D human pose estimation is often formulated as a regression problem rather than a classification task. Li and Chan~\cite{li20143d} firstly used CNNs to learn 3D human pose directly from input images. Relative 3D position to the parent joint is learned by CNNs via regression. They also used 2D part detectors of each joints in a sliding window fashion. They found that loss function which combines 2D joint classification and 3D joint regression helps to improve the 3D pose estimation results. Li et al.~\cite{li2015maximum} improved the performance of 3D pose estimation by integrating a structured learning framework into CNNs. Recently, Tekin et al.~\cite{tekin2016structured} proposed a structured prediction framework which learns 3D pose representations using an auto-encoder. Temporal information from video sequences also helps to predict more accurate pose estimation result. Zhou et al.~\cite{Zhou_2016_CVPR} used the result of 2D pose estimation to reconstruct a 3D pose. They represented a 3D pose as a weighted sum of shape bases similar to typical non-rigid structure from motion, and they designed an EM-algorithm which formulates the 3D pose as a latent variable when 2D pose estimation results are available. The method achieved the state-of-the-art performance for 3D human pose estimation when combined with 2D pose predictions learned from CNN. Tekin et al.~\cite{Tekin_2016_CVPR} used multiple consecutive frames to build a spatio-temporal features, and the features are fed to a deep neural network regressor to estimate the 3D pose.

The method proposed in this paper aims to provide an end-to-end learning framework to estimate 3D structure of a human body from a single image. Similar to~\cite{li20143d}, 3D and 2D pose information are jointly learned in a single CNN. Unlike the previous works, we directly propagate the 2D classification results to the 3D pose regressors inside the CNNs. Using additional information such as 2D classification results and the relative distance from multiple joints, we improve the performance of 3D human pose estimation over the baseline method.

\section{3D-2D Joint Estimation of Human Body Using CNN}
\label{sec:cnn}
The task of 3D human pose estimation is defined as predicting the 3D joint positions of a human body. Specifically, we estimate the relative 3D position of each joint with respect to the root joint. The number of joints $N_j$ is set to 17 in this paper according to the dataset used in the experiment. The key idea of our method is to train CNN which performs 3D pose estimation using both image features from the input image and 2D pose information retrieved from the same CNN. In other words, the proposed CNN is trained for both 2D joint classification and 3D joint regression tasks simultaneously. Details of each part is explained in the following subsections.

\begin{figure}[t]
\centering
   \includegraphics[width= 0.95\textwidth]{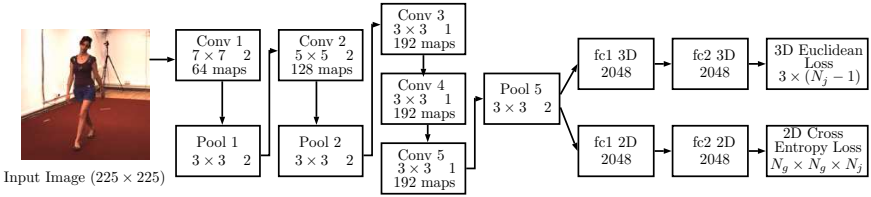}
    \caption{The baseline structure of CNN used in this paper. Convolutional and pooling layers are shared for both 2D and 3D losses, and the losses are attached to different fully connected layers.}
    \label{fig1}
\end{figure}

\subsection{Structure of the Baseline CNN}
The CNN used in this experiment consists of five convolutional layers, three pooling layers, two parallel sets of two fully connected layers, and loss layers for 2D and 3D pose estimation tasks. The CNN accepts a $225\times225$ sized image as an input. The sizes and the numbers of filters as well as the strides are specified in Figure~\ref{fig1}. The filter sizes of convolutional and pooling layers are the same as those of ZFnet~\cite{zeiler2014visualizing}, but we reduced the number of feature maps to make the network smaller.

Joint optimization using both 3D and 2D information helps CNN to learn more meaningful features than the optimization using 3D regression alone. Li et al.~\cite{li20143d} trained a CNN both for 2D joint detection task and for 3D pose regression task. Since both tasks share the same convolutional layers, features that are useful for estimating both 2D and 3D positions of joints in an image are learned in convolutional layers. Following the idea, we also used both 2D and 3D loss functions in the CNN. Convolutional layers are shared, and the feature maps after the last pooling layer are connected to two different fully connected layers, each of which is connected to 2D loss function and 3D loss function respectively (See Figure~\ref{fig1}).

We formulated 2D pose estimation as a classification problem. For the 2D classification task, we divided an input image into $N_g \times N_g$ grids and treat each grid as a separate class, which results in $N_{g}^2$ classes per joint. The ground truth label is assigned in accordance with the ground truth position of each joint. When the ground truth joint position is near the boundary of a grid, zero-one labeling that is typically used for multi-class classification may give unprecise information. Therefore, we used a soft label which assigns non-zero probability to the four nearest neighbor grids from the ground truth joint position. The target probability for the $i$th grid $g_i$ of the $j$th joint is inversely proportional to the distance from the ground truth position, i.e.,
\begin{equation}\label{eq1}
  \hat{p}_j (g_i) = \frac{d^{-1} (\mathbf{\hat{y}}_j, \mathbf{c}_i ) I( g_i )}{ \sum_{k=1}^{N_{g}^2} d^{-1} (\mathbf{\hat{y}}_j, \mathbf{c}_k ) I( g_k )} ,
\end{equation}
where $d^{-1}(\mathbf{x},\mathbf{y})$ is the inverse of the Euclidean distance between the point $\mathbf{x}$ and $\mathbf{y}$ in the 2D pixel space, $\mathbf{\hat{y}}_j$ is the ground truth position of the $j$th joint in the image, and $\mathbf{c}_i$ is the center of the grid $g_i$. $I( g_i )$ is an indicator function that is equal to 1 if the grid $g_i$ is one of the four nearest neighbors, i.e.,
\begin{equation}\label{eq1_2}
  \mathbf{I}( g_i ) = \begin{cases}
                        1 & \mbox{if }  d (\mathbf{\hat{y}}_j, \mathbf{c}_i ) < w_g \\
                        0 & \mbox{otherwise},
                      \end{cases}
\end{equation}
where $w_g$ is the width of a grid. Hence, higher probability is assigned to the grid closer to the ground truth joint position, and $\hat{p}_j (g_i)$ is normalized so that the sum of the class probabilities is equal to 1. Finally, the objective of the 2D classification task for the $j$th joint is to minimize the following cross entropy loss function.
\begin{equation}\label{eq2}
  \mathcal{L}_{2D}(j) = - \sum_{i=1}^{N_{g}^2} \hat{p}_j (g_i) \log {p_j (g_i)},
\end{equation}
where $p_j (g_i)$ is the probability that comes from the softmax output of the CNN.

On the other hand, estimating 3D position of joints is formulated as a regression task. Since the search space is much larger than the 2D case, it is undesirable to solve 3D pose estimation as a classification task. The 3D loss function is designed as a square of the Euclidean distance between the prediction and the ground truth. We estimate 3D position of each joint relative to the root node. Hence, the loss function for the $j$th joint when the root node is the $r$th joint becomes
\begin{equation}\label{eq3}
  \mathcal{L}_{3D}(j,r) = \left \lVert \mathbf{R_j} - (\mathbf{\hat{J}_j} - \mathbf{\hat{J}_{r}}) \right \rVert ^2,
\end{equation}
where $\mathbf{R_j}$ is the predicted relative 3D position of the $j$th joint from the root node, $\mathbf{\hat{J}_j}$ is the ground truth 3D position of the $j$th joint, and $\mathbf{\hat{J}_{r}}$ is that of the root node. The overall cost function of the CNN combines (\ref{eq2}) and (\ref{eq3}) with weights, i.e.,
\begin{equation}\label{eq4}
  \mathcal{L}_{all} = \lambda_{2D} \sum_{j=1}^{N_j}\mathcal{L}_{2D}(j) + \lambda_{3D} \sum_{j \neq r}^{N_j} \mathcal{L}_{3D} (j,r).
\end{equation}

\subsection{3D Joint Regression with 2D Classification Features}

\begin{figure}[t]
\centering
   \includegraphics[width= 0.99\textwidth]{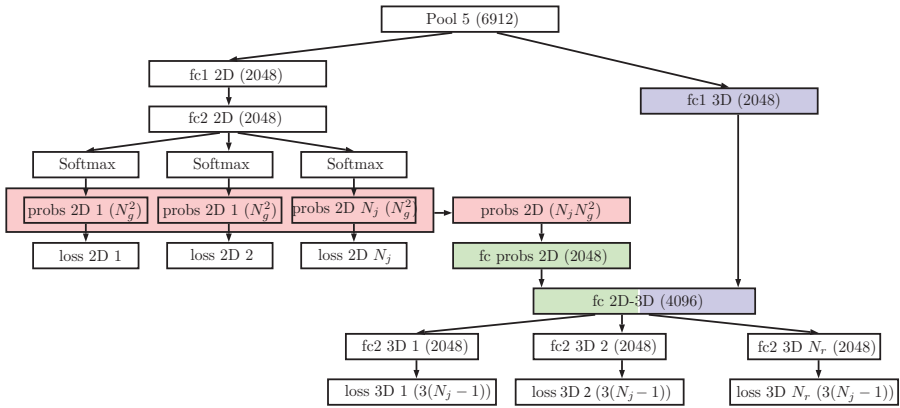}
    \caption{Structure of fully connected layers and loss functions in the proposed CNN. The numbers in parentheses indicate the dimensions of the corresponding output feature vectors.}
    \label{fig2}
\end{figure}

In the baseline architecture in Figure~\ref{fig1}, 2D and 3D losses are separated with different fully connected layers. Though convolutional layers learn features relevant to both 2D and 3D pose estimation thanks to the shared convolutional layers, the probability distribution that comes from 2D classification may give more stable and meaningful information in estimating 3D pose. The joint locations in an image are usually a strong cue for guessing 3D pose. To exploit 2D classification result as a feature for the 3D pose estimation, we concatenate the outputs of softmax in the 2D classification task with the outputs of the fully connected layers in the 3D loss part. The proposed structure after the last pooling layer is shown in Figure~\ref{fig2}. First, the 2D classification result is concatenated (\textit{probs 2D} layer in Figure~\ref{fig2}) and passes the fully connected layer (\textit{fc probs 2D}). Then, the feature vectors from 2D and 3D part are concatenated (\textit{fc 2D-3D}), which is used for 3D pose estimation task. Note that the error from the \textit{fc probs 2D} layer is not back-propagated to the \textit{probs 2D} layer to ensure that layers used for the 2D classification are trained only by the 2D loss part. The idea of using 2D classification result as an input for another task is similar to~\cite{wei2016convolutional}, which repeatedly uses the 2D classification result as an input by concatenating it with feature maps from CNN. Unlike~\cite{wei2016convolutional}, we simply vectorized the softmax result to produce $N_g \times N_g \times N_j$ feature vector rather than convolving the probability map with features in the convolutional layers.

The proposed framework can be trained end-to-end via back-propagation algorithm. Because 2D classification will give an inaccurate prediction in the early stage of training, it is possible that 3D regression may be disturbed by the classification result. However, we empirically found that 3D loss converges successfully, and the performance of 3D pose estimation improves as well, as explained in Section~\ref{sec:exp}.

\subsection{Multiple 3D Pose Regression from Different Root Nodes}
\begin{figure}[t]
\centering
  \subfigure[ ]{
  \label{fig3a}
  \includegraphics[width= 0.35\textwidth]{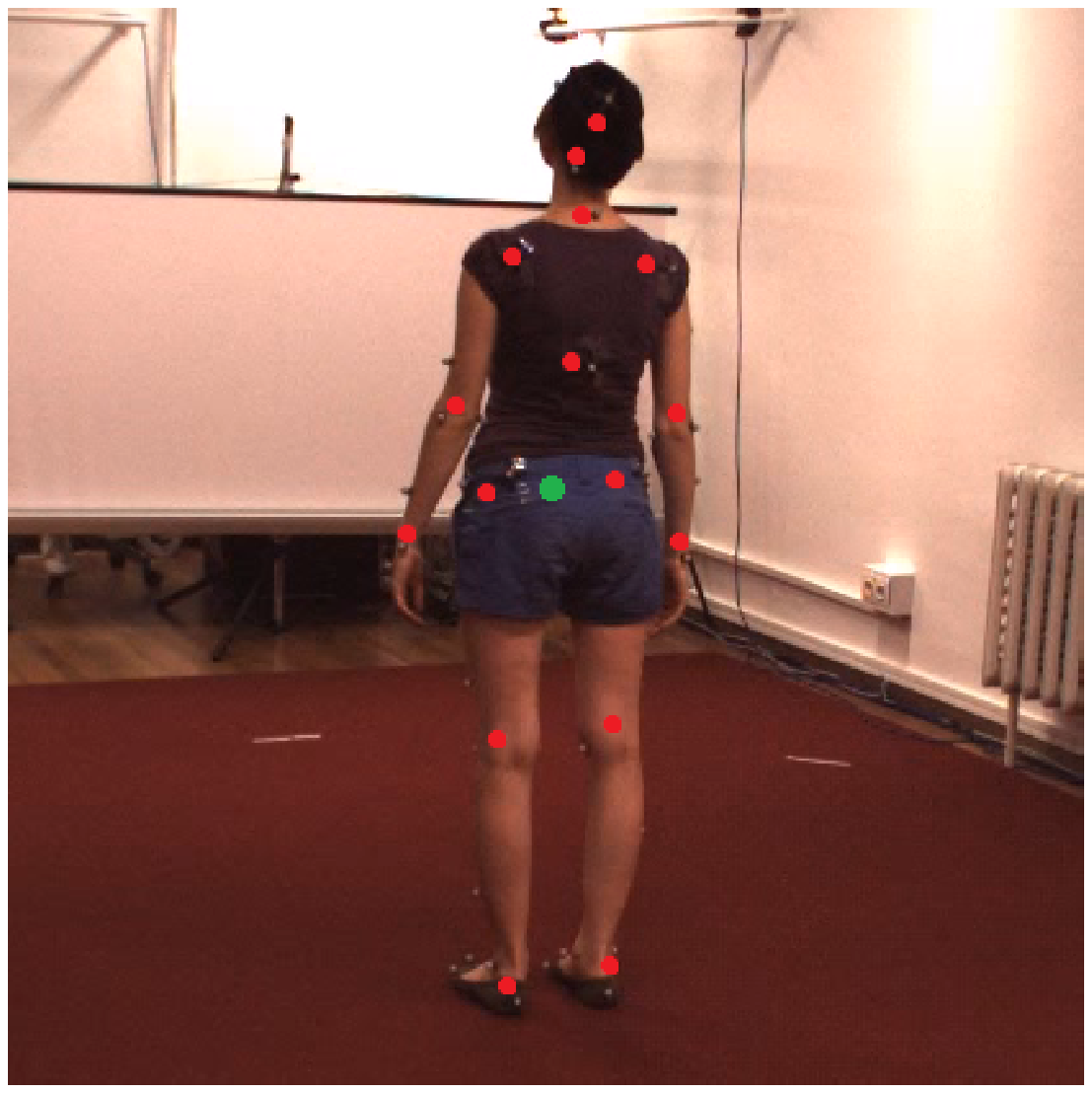}
  }
  \subfigure[ ]{
  \label{fig3b}
  \includegraphics[width= 0.35\textwidth]{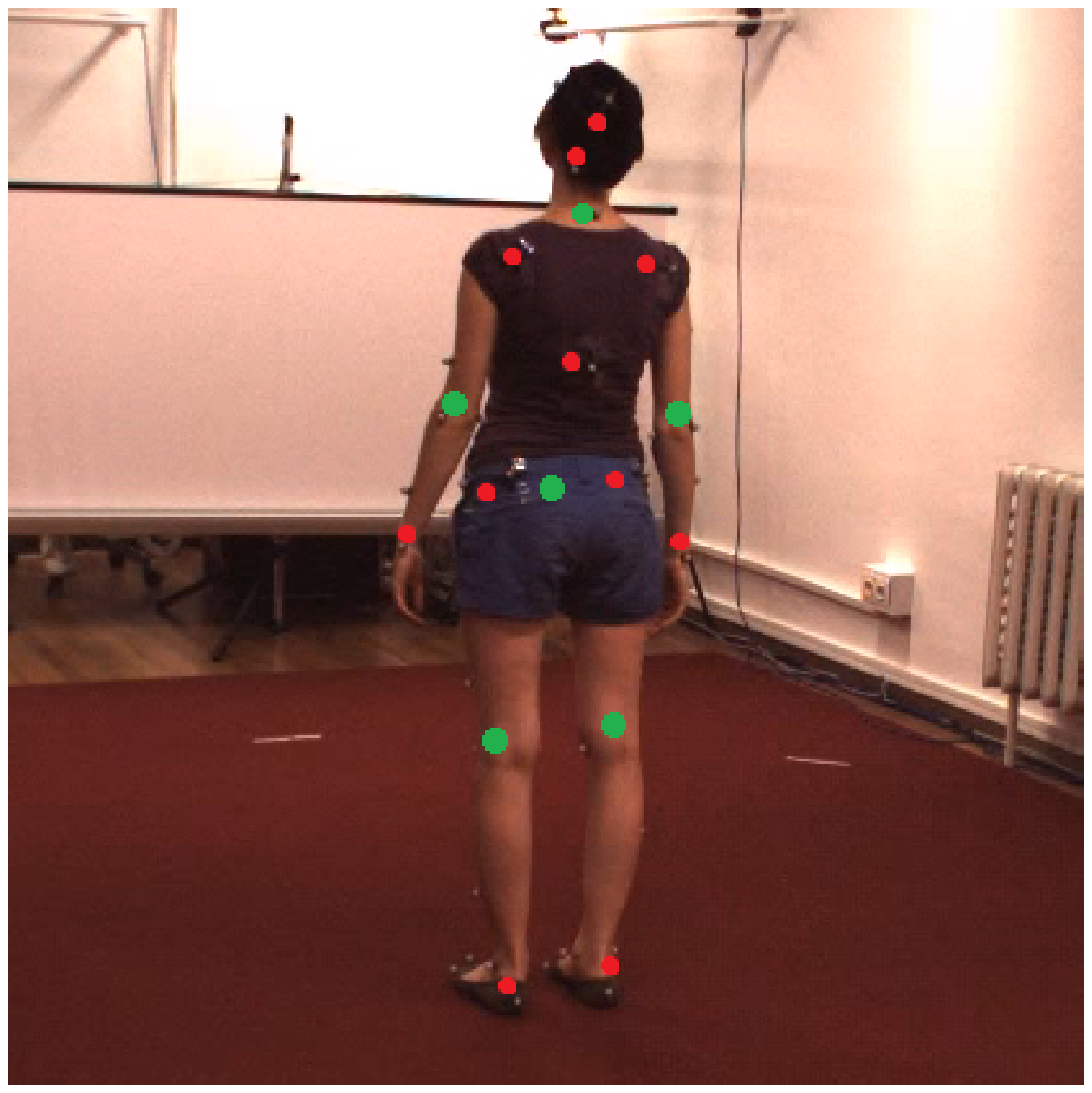}
  }
  \caption{Visualization of joints to be estimated (Red and green dots). (a) Baseline method predicts relative position of the joints with respect to one root node (Green dot). (b) For multiple pose regression, the positions of joints are estimated with respect to multiple root nodes (Green dots).}
\end{figure}

In the baseline architecture, we predicted the relative 3D position of each joint with respect to only one root node which is around the position of the hip. When joints such as wrists or ankles are far from the root node, the accuracy of regression may be degraded. Li et al.~\cite{li20143d} designed a 3D regression loss to estimate the relative position between each joint and its parent joint. However, errors may be accumulated when intermediate joint produces inaccurate result in this scheme. As an alternative solution, we estimate the relative position over multiple joints. We denote the number of selected root nodes as $N_r$. For the experiments in this paper, we set $N_r=6$ and selected six joints so that most joints can either be the root node or their neighbor nodes. The selected joints are visualized in Figure~\ref{fig3b}. Therefore, there are six 3D regression losses in the network, which is illustrated in Figure~\ref{fig2}. Then, the overall loss becomes
\begin{equation}\label{eq5}
  \mathcal{L}_{all} = \lambda_{2D} \sum_{j=1}^{N_j}\mathcal{L}_{2D}(j) + \lambda_{3D} \sum_{r \in \mathbf{R}} \sum_{j \neq r}^{N_j} \mathcal{L}_{3D} (j,r),
\end{equation}
where $\mathbf{R}$ is the set containing the joint indices that are used as root nodes. When the 3D losses share the same fully connected layers, the trained model outputs the same pose estimation results across all joints. To break this symmetry, we put the fully connected layers for each 3D losses (\textit{fc2 3D} layers in Figure~\ref{fig2}).

At the test time, all the pose estimation results are translated so that the mean of each pose becomes zero. Final prediction is generated by averaging the translated results. In other words, the 3D position of the $j$th joint $\mathbf{X}_{j}$ is calculated as
\begin{equation}\label{eq6}
  \mathbf{X}_{j} = \frac{\sum_{r \in \mathbf{R}} \mathbf{X}_{j}^{(r)}}{N_r},
\end{equation}
where $\mathbf{X}_{j}^{(r)}$ is the predicted 3D position of the $j$th joint when the $r$th joint is set to a root node.

\section{Implementation Details}
\label{sec:detail}
The proposed method is implemented using Caffe framework~\cite{jia2014caffe}. Batch normalization~\cite{ioffe2015batch} is applied to all convolutional and fully connected layers. Also, drpoout~\cite{srivastava2014dropout} is applied to every fully connected layers with drop probability of 0.3. Stochastic gradient descent of batch size 128 is used for optimization. Initial learning rate is set to 0.01, and it is decreased by a factor of 0.5 for every 4 epochs. The optimization is finished after 28 epochs. The momentum and the weight decay parameters are set to 0.9 and 0.001 respectively. The weighting parameter $\lambda_{2D}$ and $\lambda_{3D}$ are initially set to 0.1 and 0.5 respectively. $\lambda_{2D}$ is decreased to 0.01 after 16 epochs because we believe that 2D pose information plays an important role in learning informative features especially in the early stage of training.

Input images are cropped using the segmentation information provided with the dataset so that a person is located around the center of an image. The cropped image is resized to $250\times250$. We randomly cropped the resized image into an image of $225\times225$ size, then it is fed into the CNN as an input image. During the test time, only the center crop is evaluated for the pose prediction. Data augmentation based on the principal component analysis of training images~\cite{krizhevsky2012imagenet} is also applied. $N_g$ is set to 16, so the input image is divided into 256 square grids for 2D loss calculation. $N_r$ is set to 6, and the position of the root nodes are illustrated in Figure~\ref{fig3b}.

For the ground truth 3D pose that is used in the training step, we firstly translated the joints to make the shape to be zero mean. Then, we scaled the 3D shape so that the Frobenius norm of the 3D shape becomes 1. Since different person has different height and size, we believe that the normalization helps to reduce ambiguity of scale and to predict scale-invariant poses. During the testing phase, scale should be recovered to evaluate the performance of the algorithm. Similar to~\cite{Zhou_2016_CVPR}, we infer the scale using the training data. The lengths of all connected joints from the training set are averaged. The scale of the result from the test data is determined so that the length of connected joints in the estimated shape is equal to the pre-calculated average length. Since the lengths for arms and legs from the estimated shape often have a large variation, we only used the length of joints in the torso which is stable in most cases.

\section{Experimental Results}
\label{sec:exp}
We used Human 3.6m dataset~\cite{h36m_pami} to evaluate our method and compared the proposed method with the other 3D human pose estimation algorithms. The dataset provides 3D human pose information acquired by a motion capture system with synchronized RGB images. It consists of 15 different sequences which contain specific actions such as discussion, eating, walking, and so on. There are 7 different persons who perform all 15 actions. We trained and tested each action individually. Following the previous works on the dataset~\cite{li20143d,Zhou_2016_CVPR}, we used 5 subjects (S1, S5, S6, S7, S8) as a training set, and 2 subjects (S9, S11) as a test set. The training and the testing procedures are conducted on a single PC with a Titan X GPU. Training procedure takes 7-10 hours for one action sequence depending on the number of training images. For the evaluation metric, we used the mean per joint position error (MPJPE).

\begin{table}[t]
\centering
\caption{Quantitative results on Human 3.6m dataset. The best and the second best methods for each sequence are marked as (1) and (2) respectively.}
\label{tab1}
\begin{tabular}{ l c c c c c c }
  \hline
  \quad & Directions & Discussion & Eating & Greeting & Phoning & Photo \\ \hline
  LinKDE~\cite{h36m_pami}\quad & 132.71 & 183.55 & 132.37 & 164.39 & 162.12 & 205.94 \\
  Li and Chan~\cite{li20143d}\quad & - & 148.79 & 104.01 & 127.17 & - & 189.08 \\
  Li et al.~\cite{li2015maximum}\quad & - & 136.88 & 96.94 & 124.74 & - & 168.68 \\
  Tekin et al.~\cite{tekin2016structured}\quad & - & 129.06 & 91.43 & 121.68 & - & 162.17 \\
  Tekin et al.~\cite{Tekin_2016_CVPR}\quad & 102.41 & 147.72 & \bf 88.83$^{(2)}$ & 125.28 & 118.02 & 182.73 \\
  Zhou et al.~\cite{Zhou_2016_CVPR}\quad & \bf 87.36$^{(1)}$ & \bf 109.31$^{(1)}$ & \bf 87.05$^{(1)}$ & \bf 103.16$^{(1)}$ & \bf 116.18$^{(2)}$ & \bf 143.32$^{(1)}$ \\
  Our method\quad & \bf 100.34$^{(2)}$ & \bf 116.19$^{(2)}$ & 89.96 & \bf 116.49$^{(2)}$ & \bf 115.34$^{(1)}$ & \bf 149.55$^{(2)}$ \\ \hline

  \multirow{2}{*}{\quad} & \multirow{2}{*}{Posing} & \multirow{2}{*}{Purchases} & \multirow{2}{*}{Sitting} & Sitting & \multirow{2}{*}{Smoking} & \multirow{2}{*}{Waiting} \\
  & & & & Down & & \\ \hline
  LinKDE~\cite{h36m_pami}\quad & 150.61 & 171.31 & 151.57 & 243.03 & 162.14 & 170.69 \\
  Li and Chan~\cite{li20143d}\quad & - & - & - & - & - & - \\
  Li et al.~\cite{li2015maximum}\quad & - & - & - & - & - & - \\
  Tekin et al.~\cite{tekin2016structured}\quad & - & - & - & - & - & - \\
  Tekin et al.~\cite{Tekin_2016_CVPR}\quad & \bf 112.38$^{(2)}$ & 129.17 & 138.89 & 224.90 & 118.42 & 138.75 \\
  Zhou et al.~\cite{Zhou_2016_CVPR}\quad & \bf 106.88$^{(1)}$ & \bf 99.78$^{(1)}$ & \bf 124.52$^{(1)}$ & \bf 199.23$^{(2)}$ & \bf 107.42$^{(2)}$ & \bf 118.09$^{(1)}$ \\
  Our method\quad & \quad117.57 & \bf 106.94$^{(2)}$ & \bf 137.21$^{(2)}$ & \bf 190.82$^{(1)}$ & \bf 105.78$^{(1)}$ & \bf 125.12$^{(2)}$ \\ \hline

  \multirow{2}{*}{\quad} & Walk & \multirow{2}{*}{Walking} & Walk & \multirow{2}{*}{Average} \\
  & Dog & & Together & \\ \hline
  LinKDE~\cite{h36m_pami}\quad & 177.13 & 96.60 & 127.88 & 162.14 \\
  Li and Chan~\cite{li20143d}\quad & 146.59 & 77.60 & - & - \\
  Li et al.~\cite{li2015maximum}\quad & 132.17 & 69.97 & - & - \\
  Tekin et al.~\cite{tekin2016structured}\quad & 130.53 & 65.75 & - & - \\
  Tekin et al.~\cite{Tekin_2016_CVPR}\quad & \bf 126.29$^{(2)}$ & \bf 55.07$^{(1)}$ & \bf 65.76$^{(1)}$ & 124.97 \\
  Zhou et al.~\cite{Zhou_2016_CVPR}\quad & \bf 114.23$^{(1)}$ & 79.39 & 97.70 & \bf 113.01$^{(1)}$ \\
  Our method\quad & 131.90 & \bf 62.64$^{(2)}$ & \bf 96.18$^{(2)}$ & \bf 117.34$^{(2)}$ \\ \hline

\end{tabular}
\end{table}

First, we compared the performance of our method with the conventional methods on Human 3.6m dataset. Table~\ref{tab1} shows the MPJPE of our method and the previous works. The smallest and the second smallest errors for each sequence are marked. Our method achieves the best performance in 3 sequences and shows the second best performance in 9 sequences. Note that the methods of \cite{Tekin_2016_CVPR} and \cite{Zhou_2016_CVPR} make use of temporal information from multiple frames. Meanwhile, our method produce a 3D pose from a single image. Our method is also beneficial against \cite{Tekin_2016_CVPR} and \cite{Zhou_2016_CVPR} in terms of running time and the simplicity of the algorithm since the estimation is done by a forward pass of the CNN and simple averaging. Moreover, from Table~\ref{tab1}, it is justified that our method outperforms the CNN based methods that predict 3D pose from a single image~\cite{li20143d,li2015maximum,tekin2016structured}.

Next, we measured the effect of our contribution, 1) the integration of 2D classification results and 2) regression from multiple root nodes, by comparing their performance with the baseline CNN. Note that the 2D classification loss is also used in the baseline CNN. The difference of the baseline CNN is that 2D classification results are not propagated to the 3D loss part, i.e., \textit{probs 2D}, \textit{fc probs 2D} and \textit{fc 2D-3D} layers in Figure~\ref{fig2} are deleted in the baseline CNN. The results are shown in Table~\ref{tab2}. Multiple regression from different root nodes and the integration of 2D classification results are denoted as \textit{Multi-reg} and \textit{2D-cls} respectively. Both modifications improve the result over the baseline CNN in all tested sequences. 2D classification integration showed larger error reduction rate than the multiple regression strategy, which proves that the 2D classification information is indeed a useful feature for 3D pose estimation. Multiple regression can be considered as an ensemble of different estimation results, which improves the overall performance. It can be found that the error reduction rate for the case that both 2D classification result integration and multiple regression are applied is slightly bigger than the sum of the reduction rates when they are individually applied in most sequences. Since each 3D pose regressor takes advantage of 2D classification feature, there is a synergy effect between the two schemes.

\begin{table}[t]
\centering
\caption{Comparison of our method with the baseline.}
\label{tab2}
\begin{tabular}{ l c c c c c c }
  \hline
  \quad & \, Discussion \,  &  \, Eating \,  &  \, Greeting \,  &  \, Phoning \,  &  \, Photo \,  &  \, Walking \,  \\ \hline
  Baseline CNN & 125.45 & 95.21 & 120.69 & 119.66 & 153.76 & 72.55 \\
  Multi-reg & 122.71 & 94.67 & 119.70 & 119.25 & 153.54 & 71.19 \\
  2D-cls & 118.19 & 91.39 & 118.19 & 115.84 & 149.97 & 64.27 \\
  Multi-reg+2D-cls & \bf 116.19 & \bf 89.96 & \bf 116.49 & \bf 115.34 & \bf 149.55 & \bf 62.64 \\ \hline
\end{tabular}
\end{table}

\begin{figure}[t]
\centering
  \subfigure[ ]{
  \label{fig4a}
  \includegraphics[width= 0.4\textwidth]{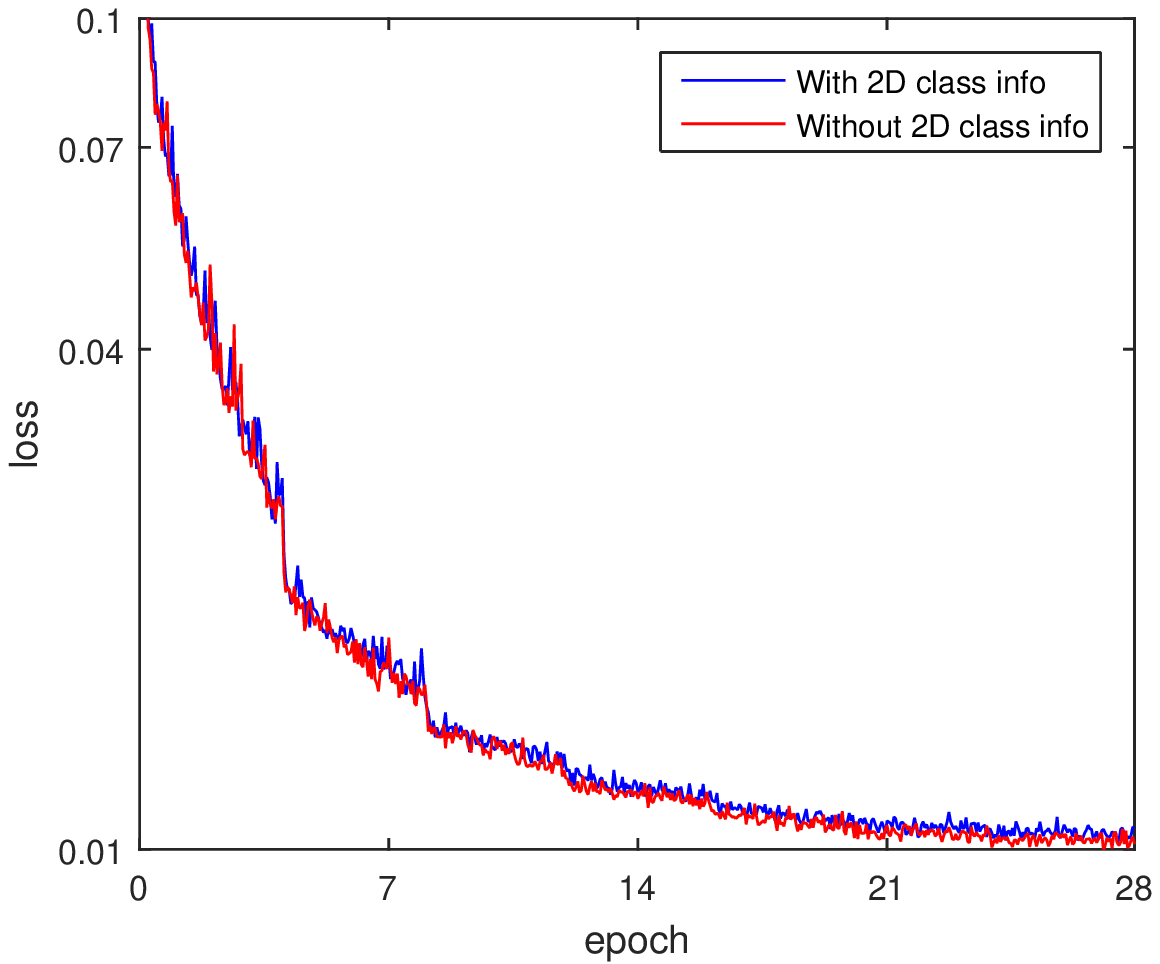}
  }
  \subfigure[ ]{
  \label{fig4b}
  \includegraphics[width= 0.4\textwidth]{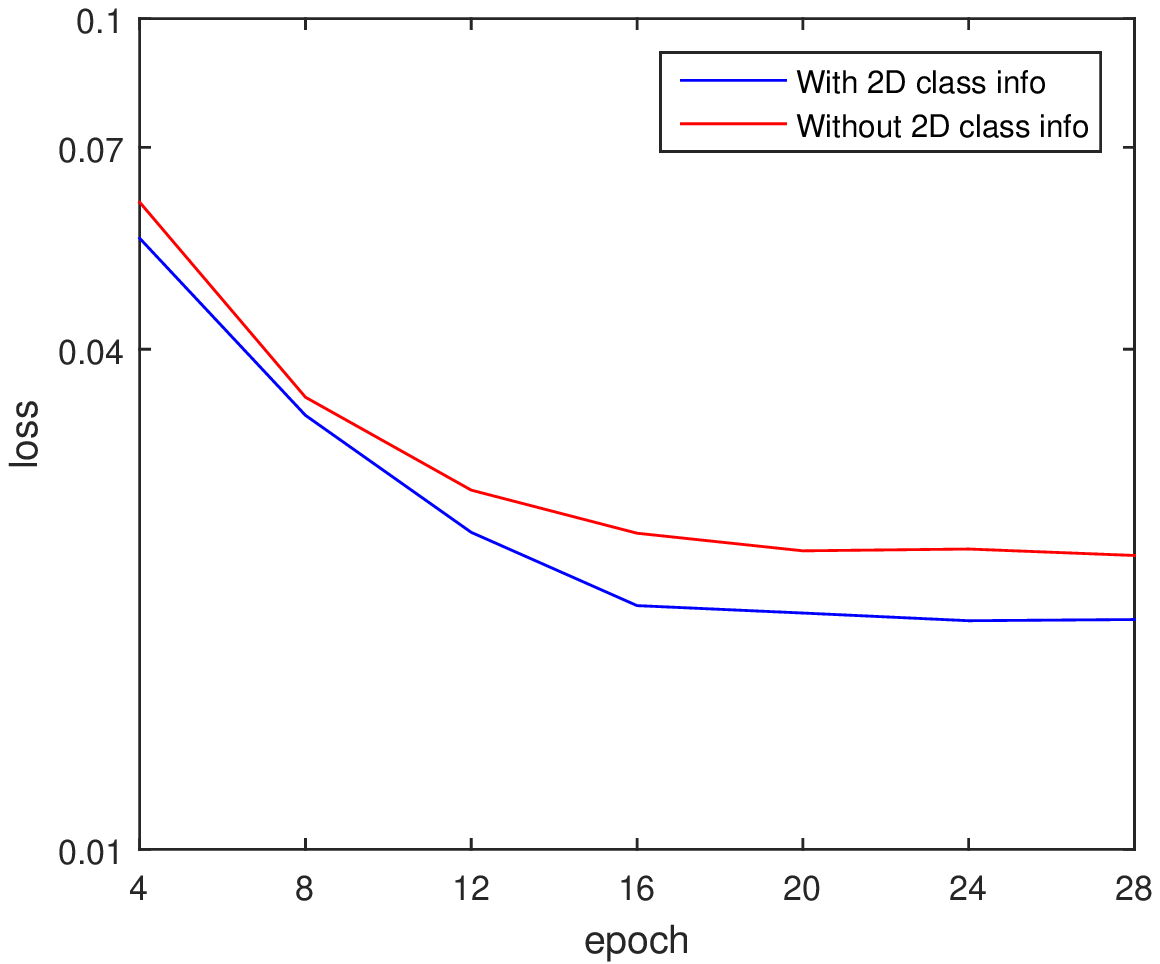}
  }
  \caption{The 3D losses of Walking sequence with and without 2D classification result integration. (a) Losses for training data. (b) Losses for test data.}
  \label{fig4}
\end{figure}

We also analyzed the effect of integrating 2D classification result in terms of 3D losses. Training losses are measured every 50 iterations and testing losses are measured every 4 epochs. The results on the Walking sequence are illustrated in Figure~\ref{fig4}. For the training data, loss is slightly smaller when 2D classification information is not used (Figure~\ref{fig4a}). However, test loss is much lower when 2D classification information is used(Figure~\ref{fig4b}). This indicates that 2D classification information impose generalization power and reduce overfitting for CNN regressor. Since the 2D joint probabilities provide more abstracted and subject-independent information compared to the features obtained from an image, the CNN model is able to learn representations that are robust to variability of subjects in the image.

\begin{figure}[p]
\centering
    \begin{minipage}{.22\textwidth}
        \centering
        \includegraphics[width=0.99\textwidth]{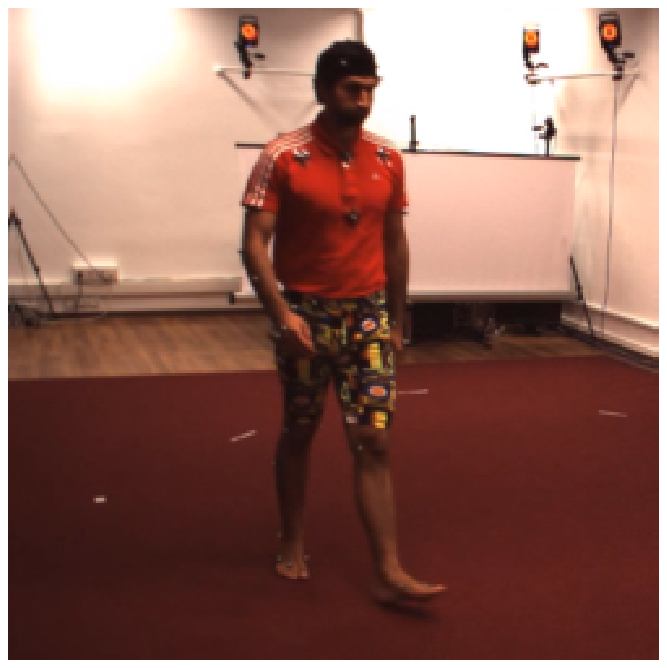}
    \end{minipage}%
    \begin{minipage}{0.25\textwidth}
        \centering
        \includegraphics[width=0.99\textwidth]{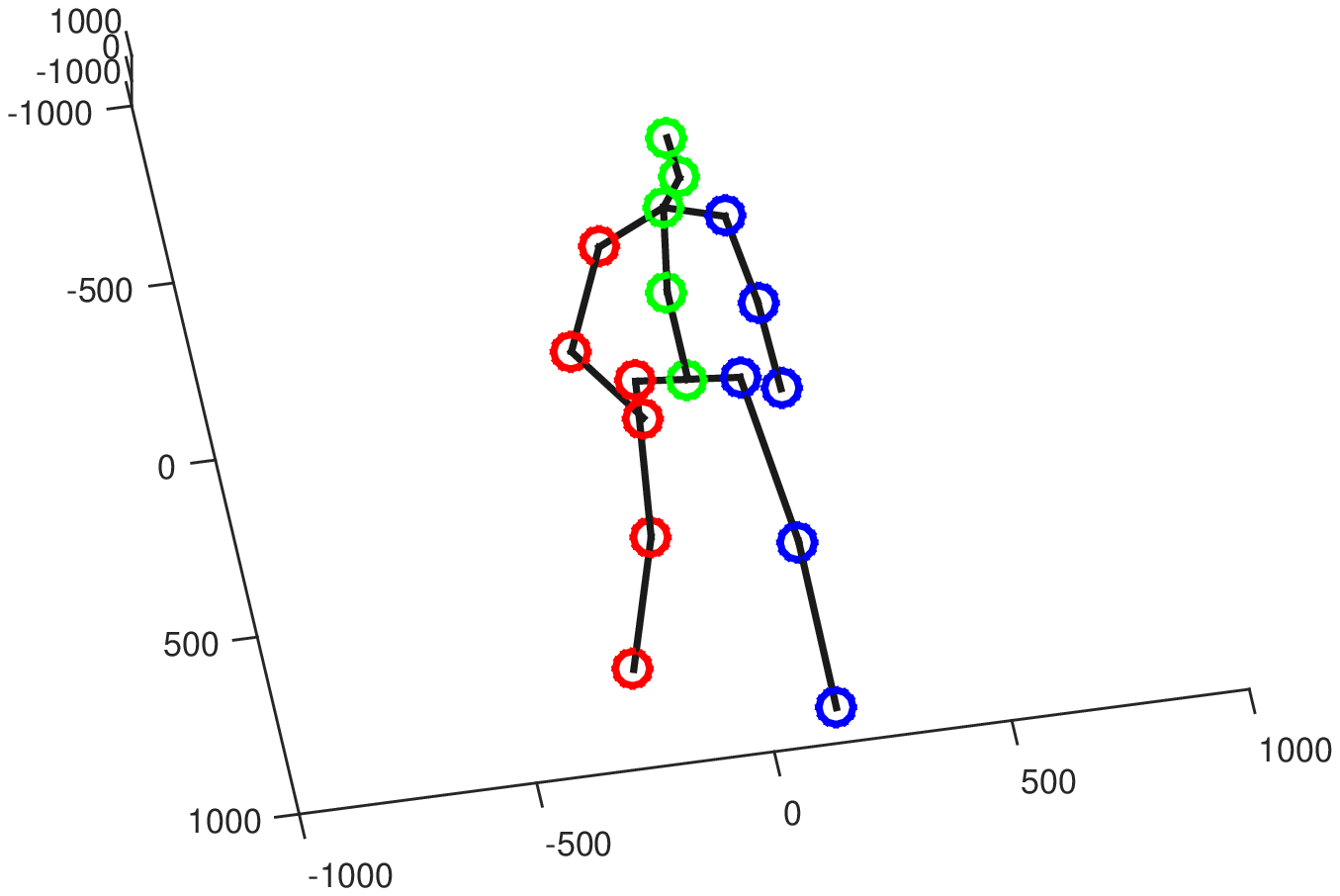}
    \end{minipage}
    \begin{minipage}{0.25\textwidth}
        \centering
        \includegraphics[width=0.99\textwidth]{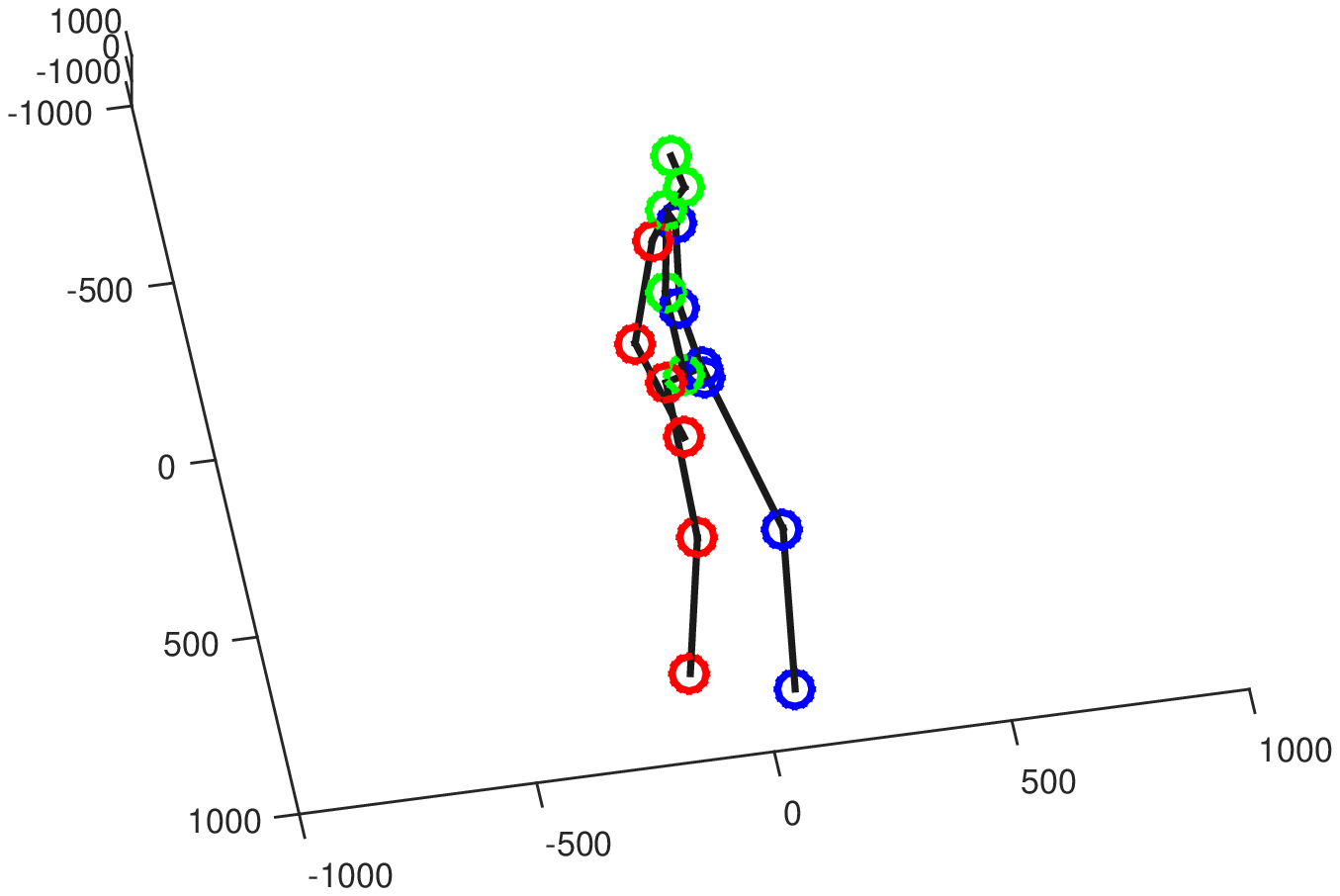}
    \end{minipage}
    \begin{minipage}{0.25\textwidth}
        \centering
        \includegraphics[width=0.99\textwidth]{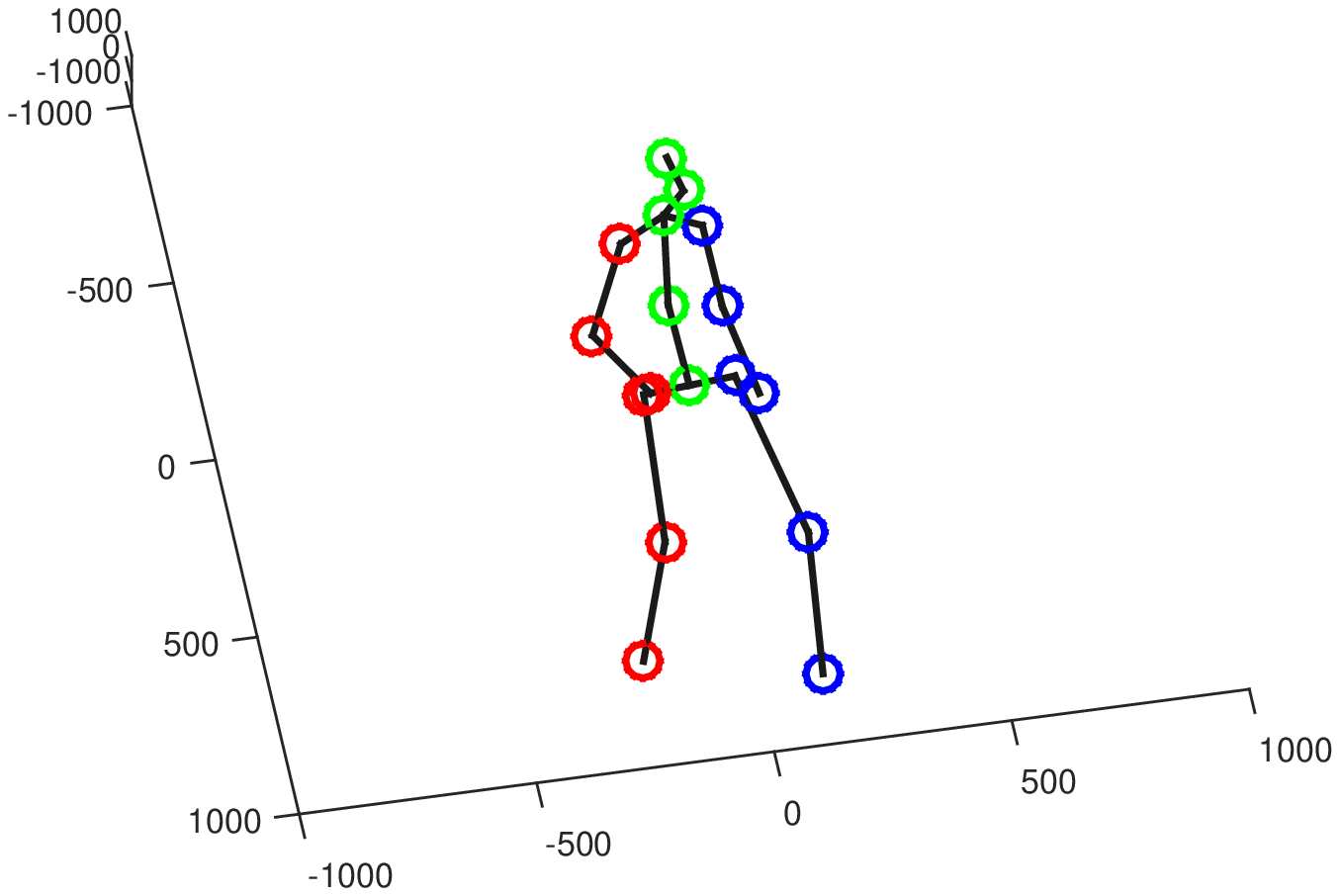}
    \end{minipage}

    \begin{minipage}{.22\textwidth}
        \centering
        \includegraphics[width=0.99\textwidth]{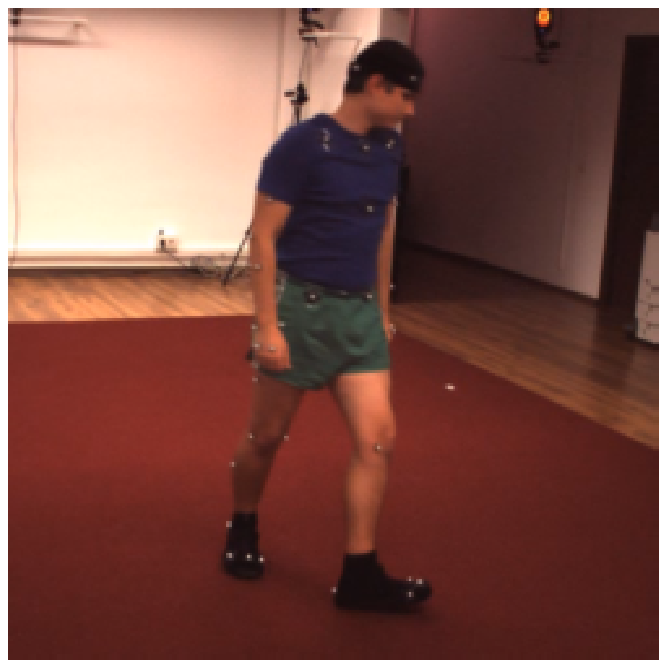}
    \end{minipage}%
    \begin{minipage}{0.25\textwidth}
        \centering
        \includegraphics[width=0.99\textwidth]{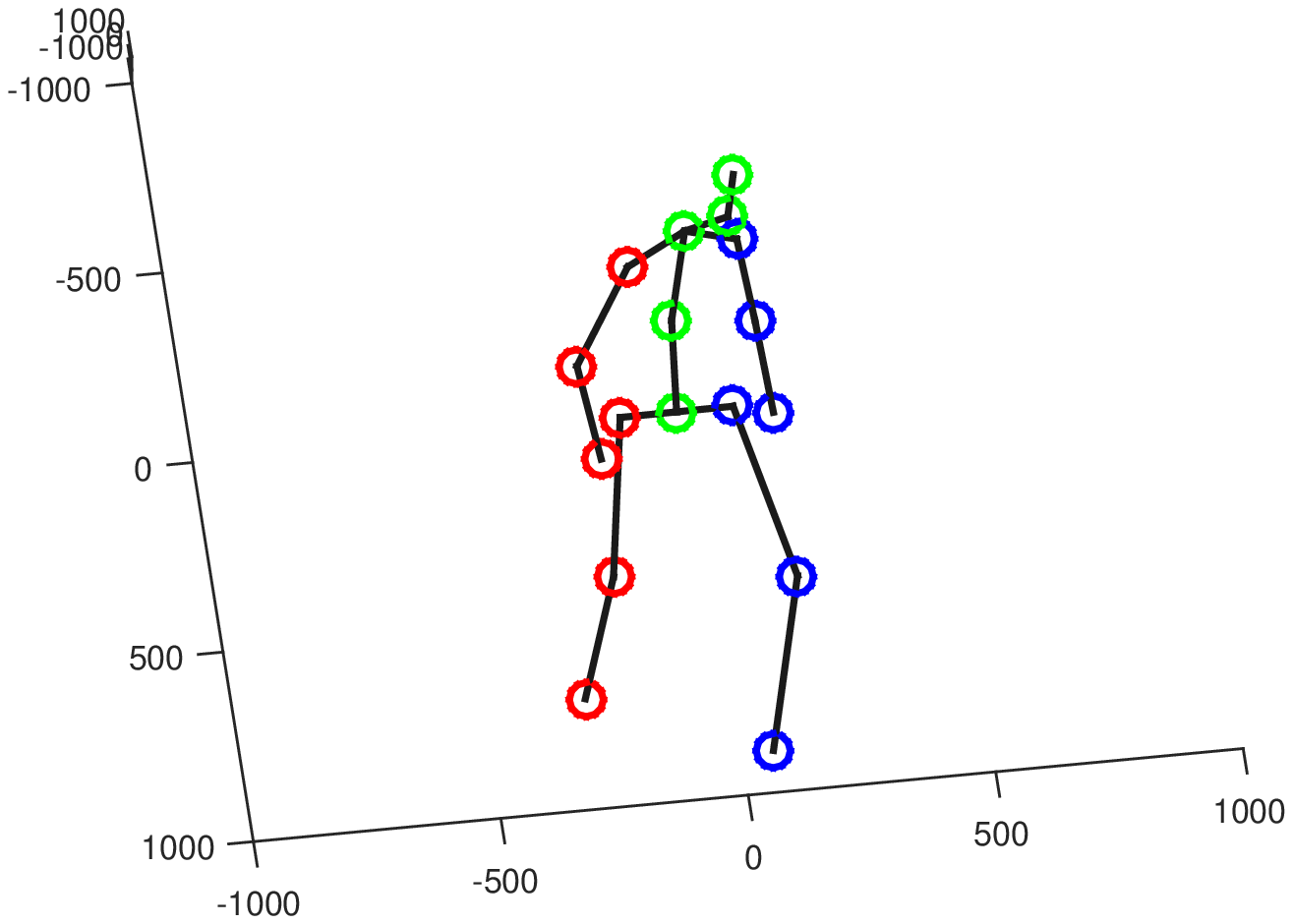}
    \end{minipage}
    \begin{minipage}{0.25\textwidth}
        \centering
        \includegraphics[width=0.99\textwidth]{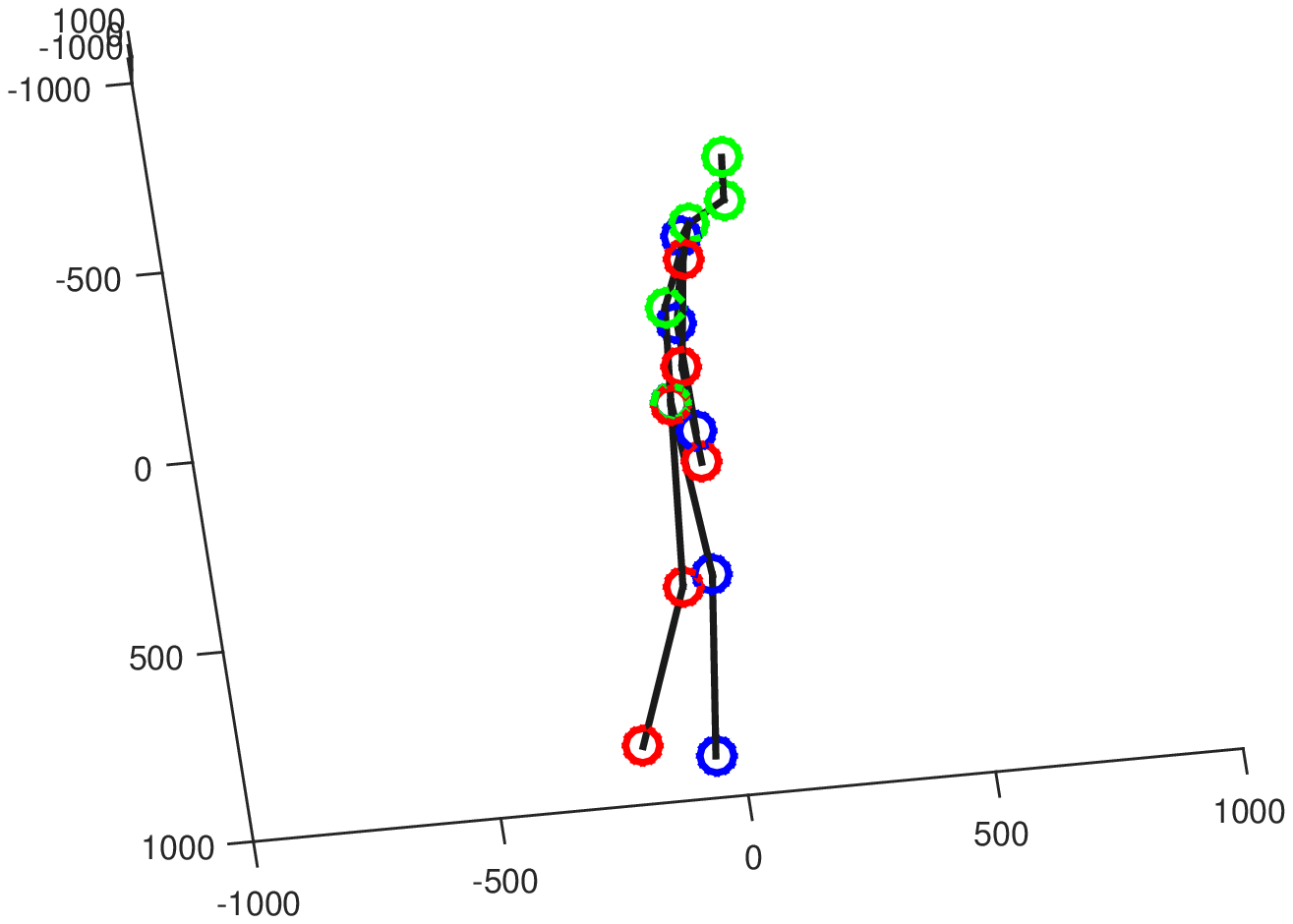}
    \end{minipage}
    \begin{minipage}{0.25\textwidth}
        \centering
        \includegraphics[width=0.99\textwidth]{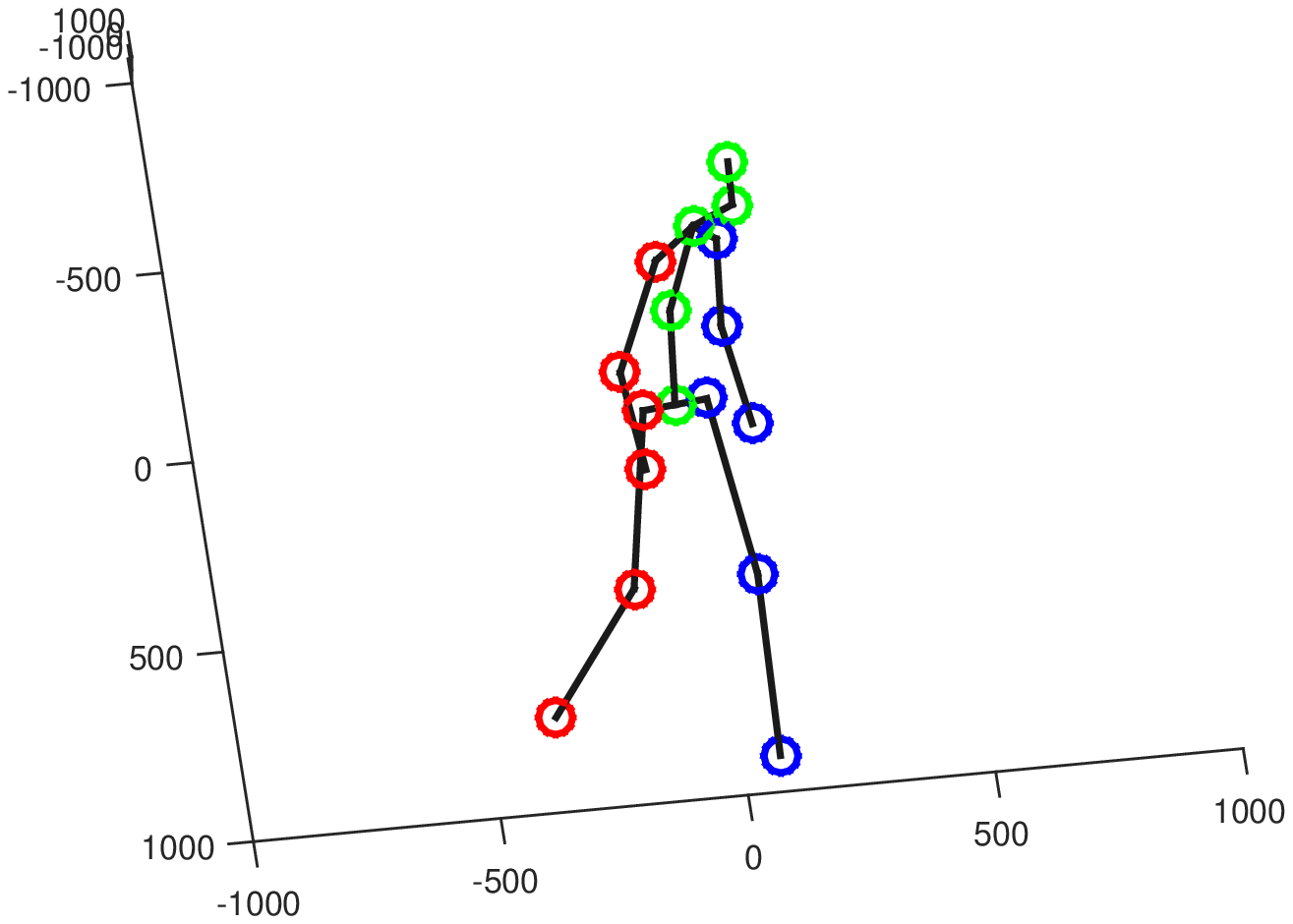}
    \end{minipage}

    \begin{minipage}{.22\textwidth}
        \centering
        \includegraphics[width=0.99\textwidth]{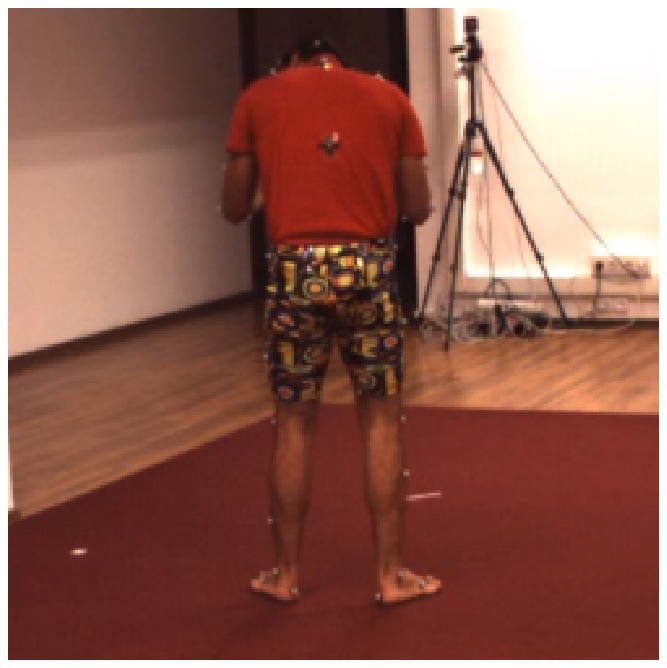}
    \end{minipage}%
    \begin{minipage}{0.25\textwidth}
        \centering
        \includegraphics[width=0.99\textwidth]{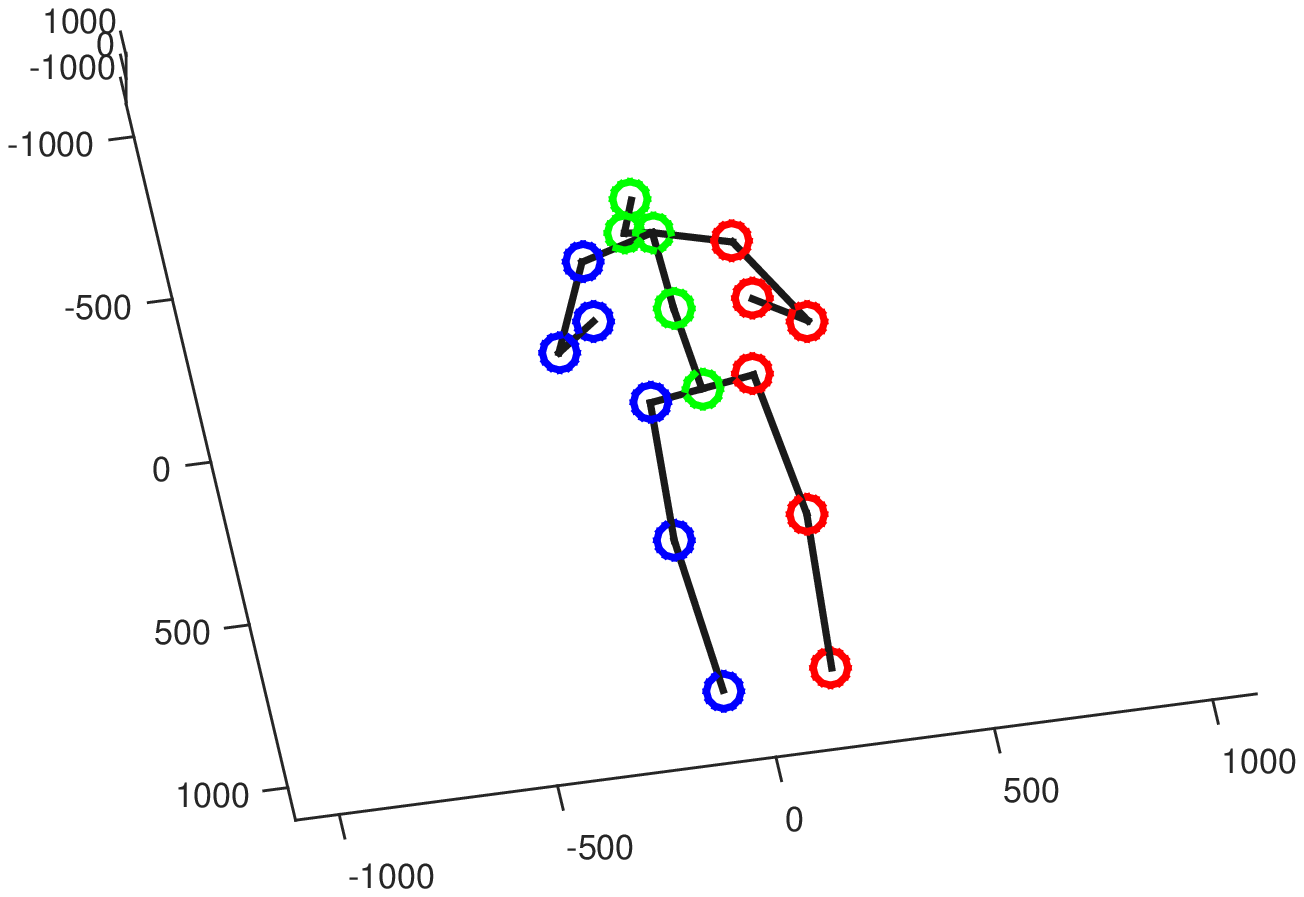}
    \end{minipage}
    \begin{minipage}{0.25\textwidth}
        \centering
        \includegraphics[width=0.99\textwidth]{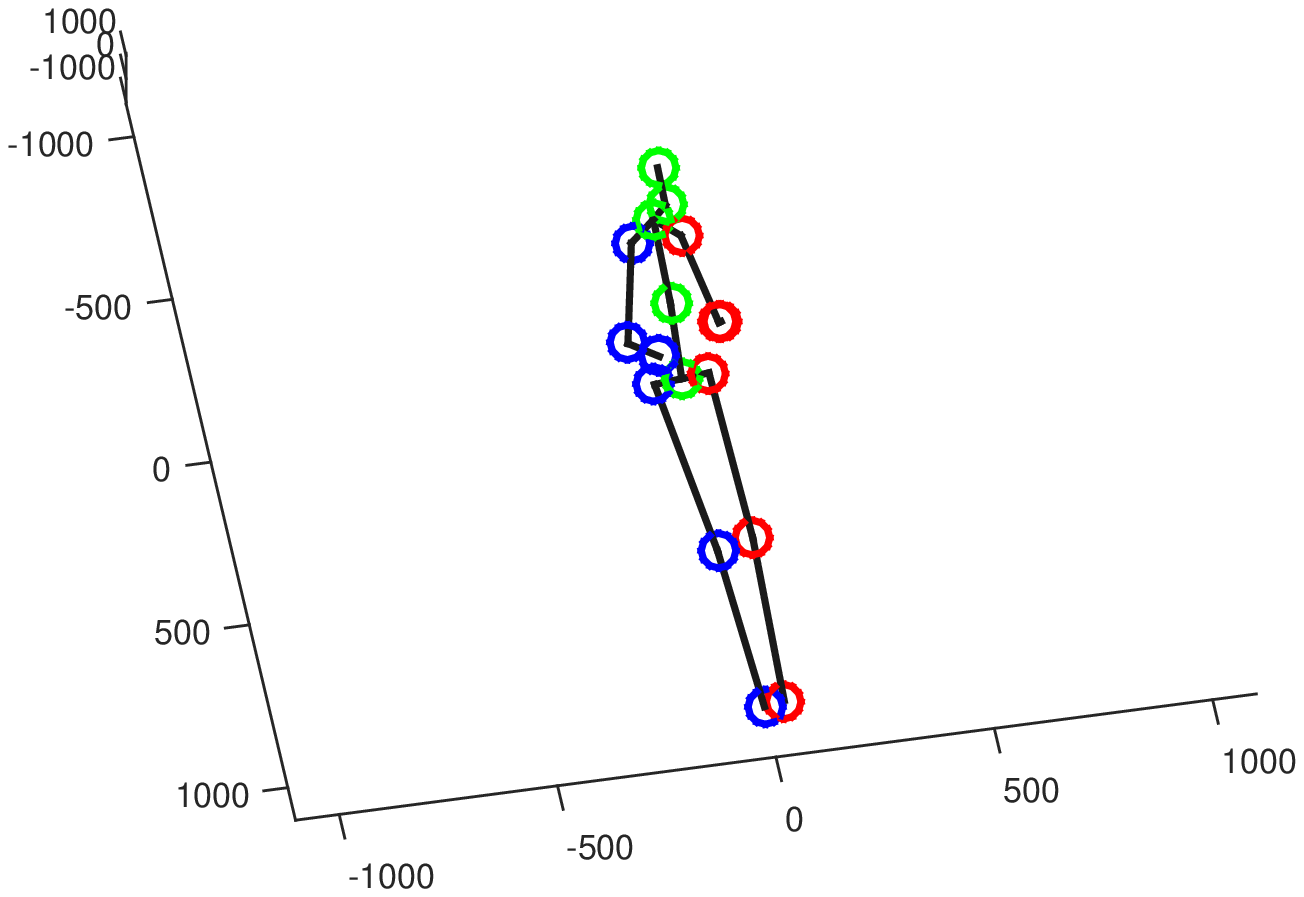}
    \end{minipage}
    \begin{minipage}{0.25\textwidth}
        \centering
        \includegraphics[width=0.99\textwidth]{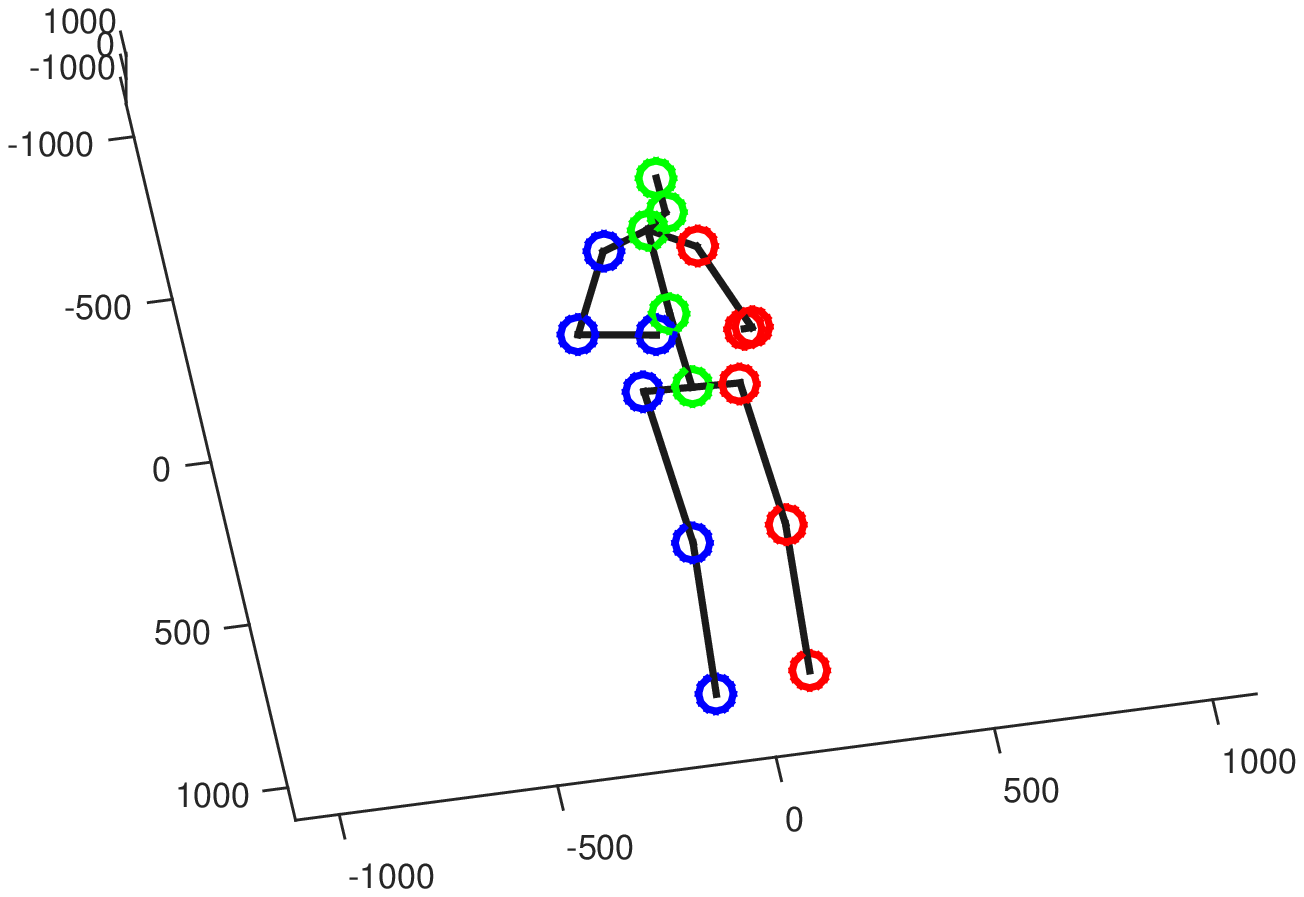}
    \end{minipage}

    \begin{minipage}{.22\textwidth}
        \centering
        \includegraphics[width=0.99\textwidth]{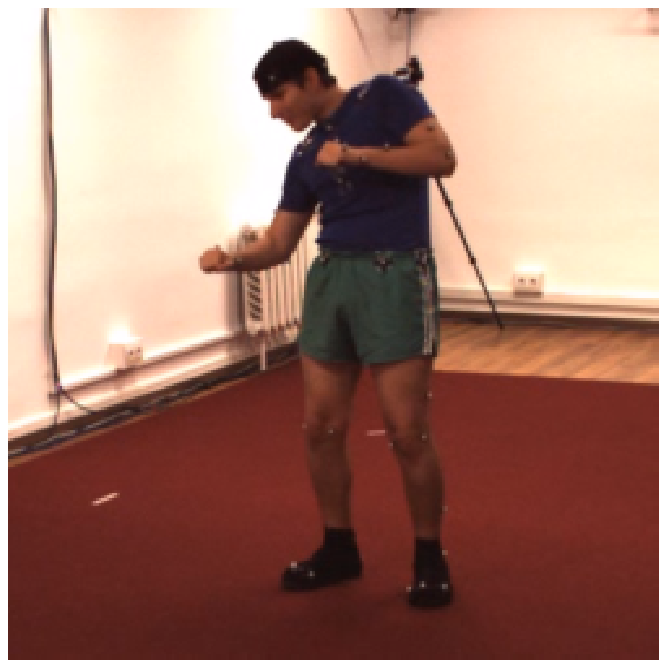}
    \end{minipage}%
    \begin{minipage}{0.25\textwidth}
        \centering
        \includegraphics[width=0.99\textwidth]{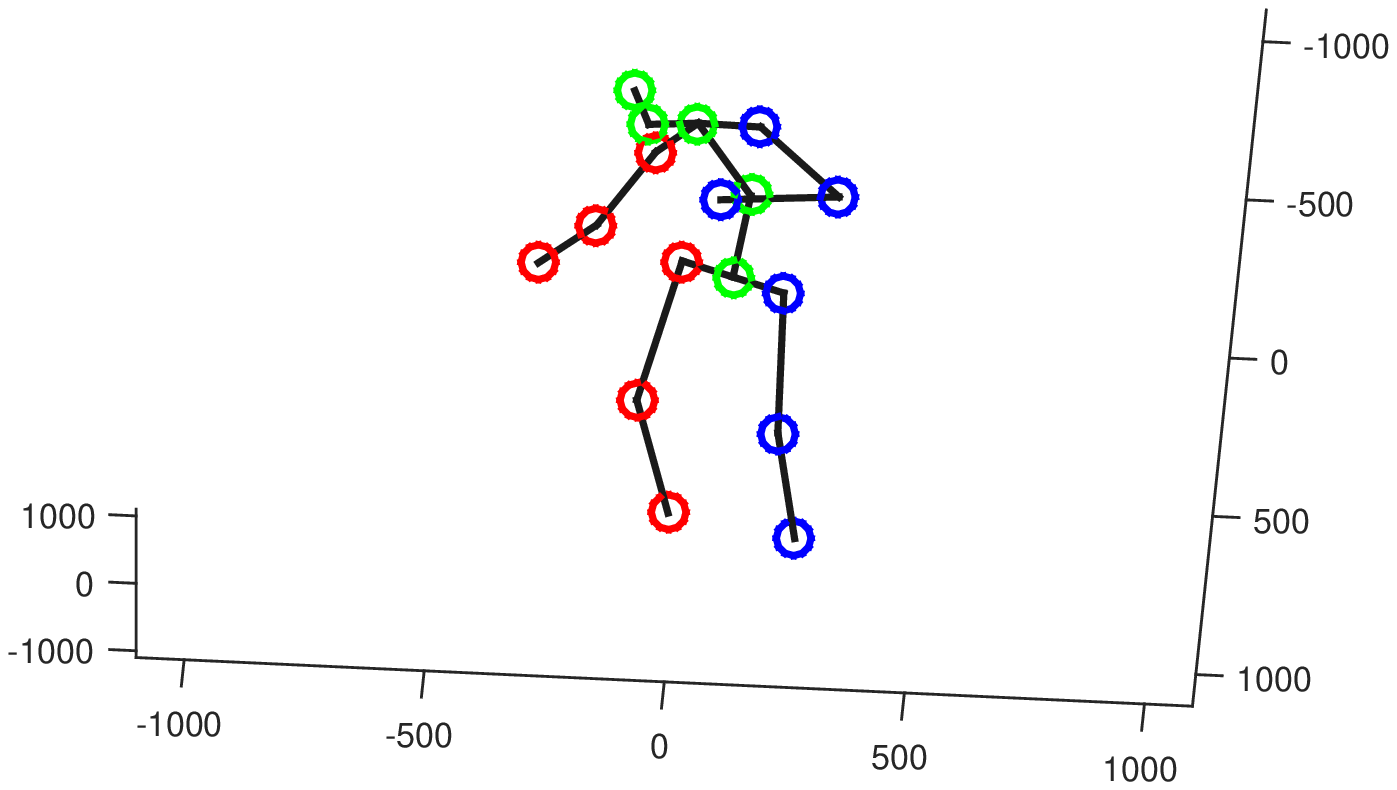}
    \end{minipage}
    \begin{minipage}{0.25\textwidth}
        \centering
        \includegraphics[width=0.99\textwidth]{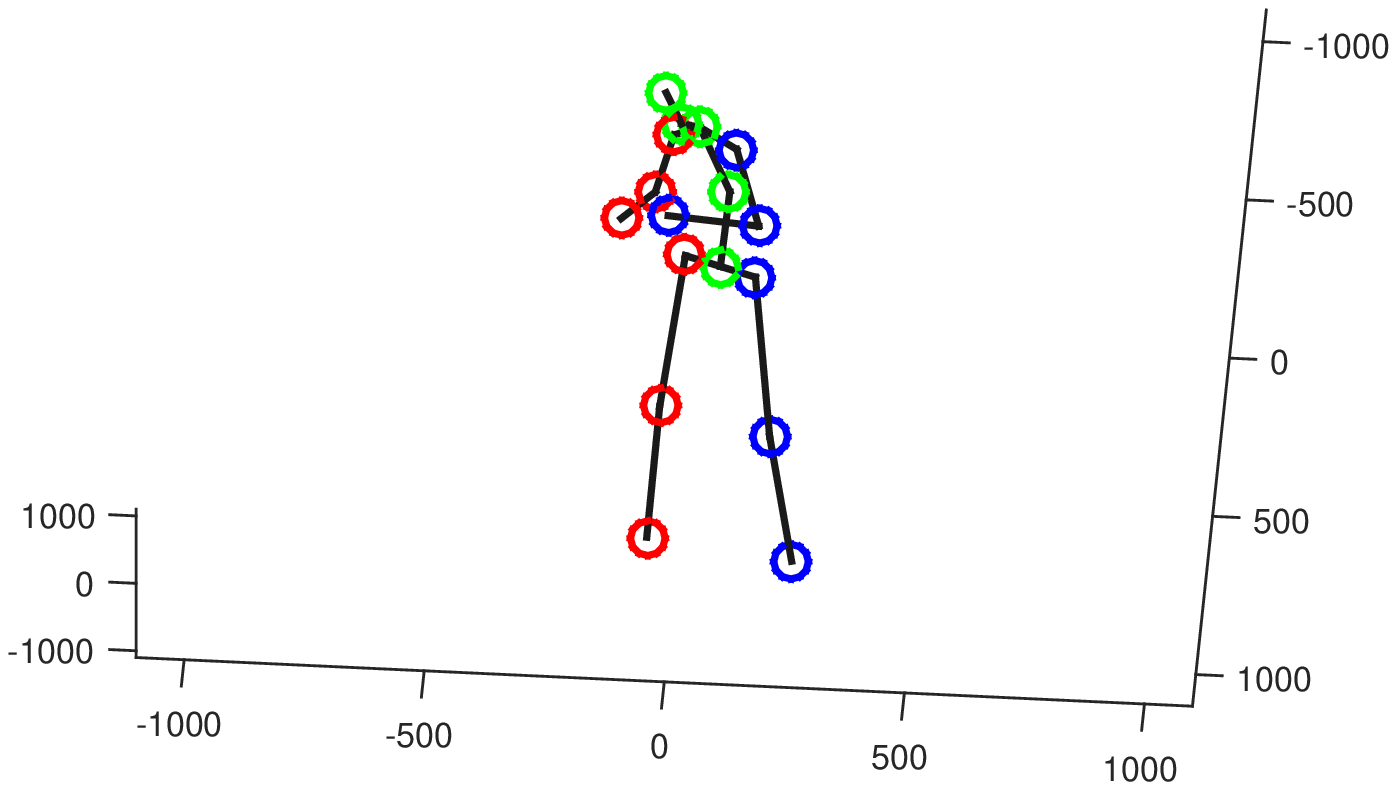}
    \end{minipage}
    \begin{minipage}{0.25\textwidth}
        \centering
        \includegraphics[width=0.99\textwidth]{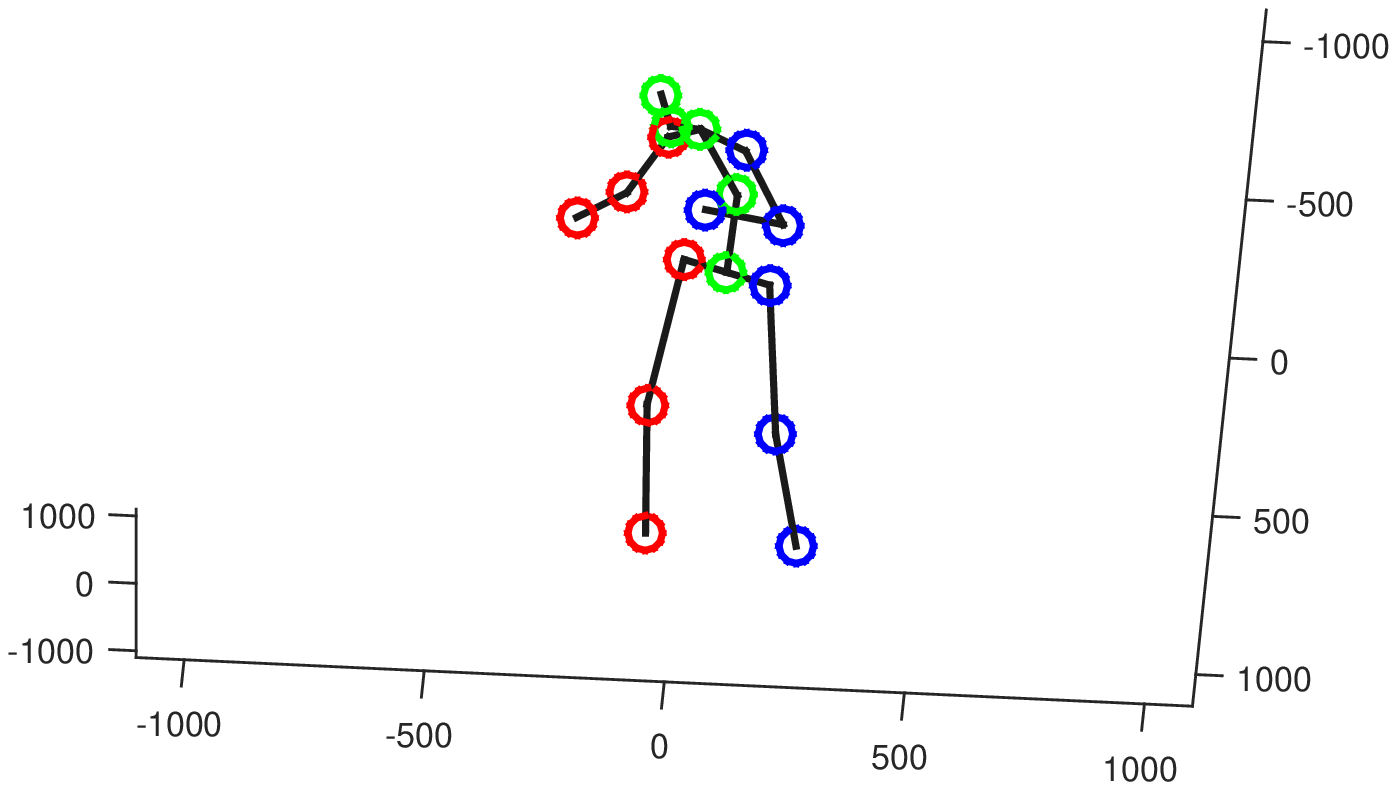}
    \end{minipage}

    \begin{minipage}{.22\textwidth}
        \centering
        \includegraphics[width=0.99\textwidth]{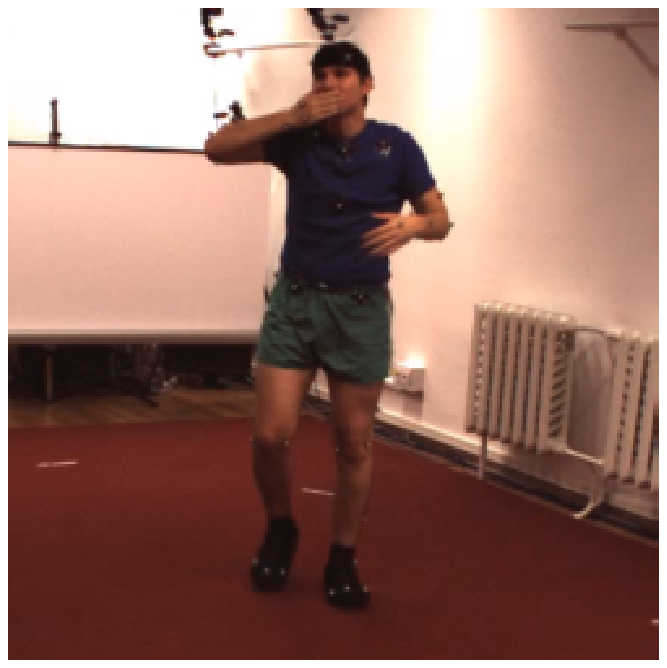}
    \end{minipage}%
    \begin{minipage}{0.25\textwidth}
        \centering
        \includegraphics[width=0.99\textwidth]{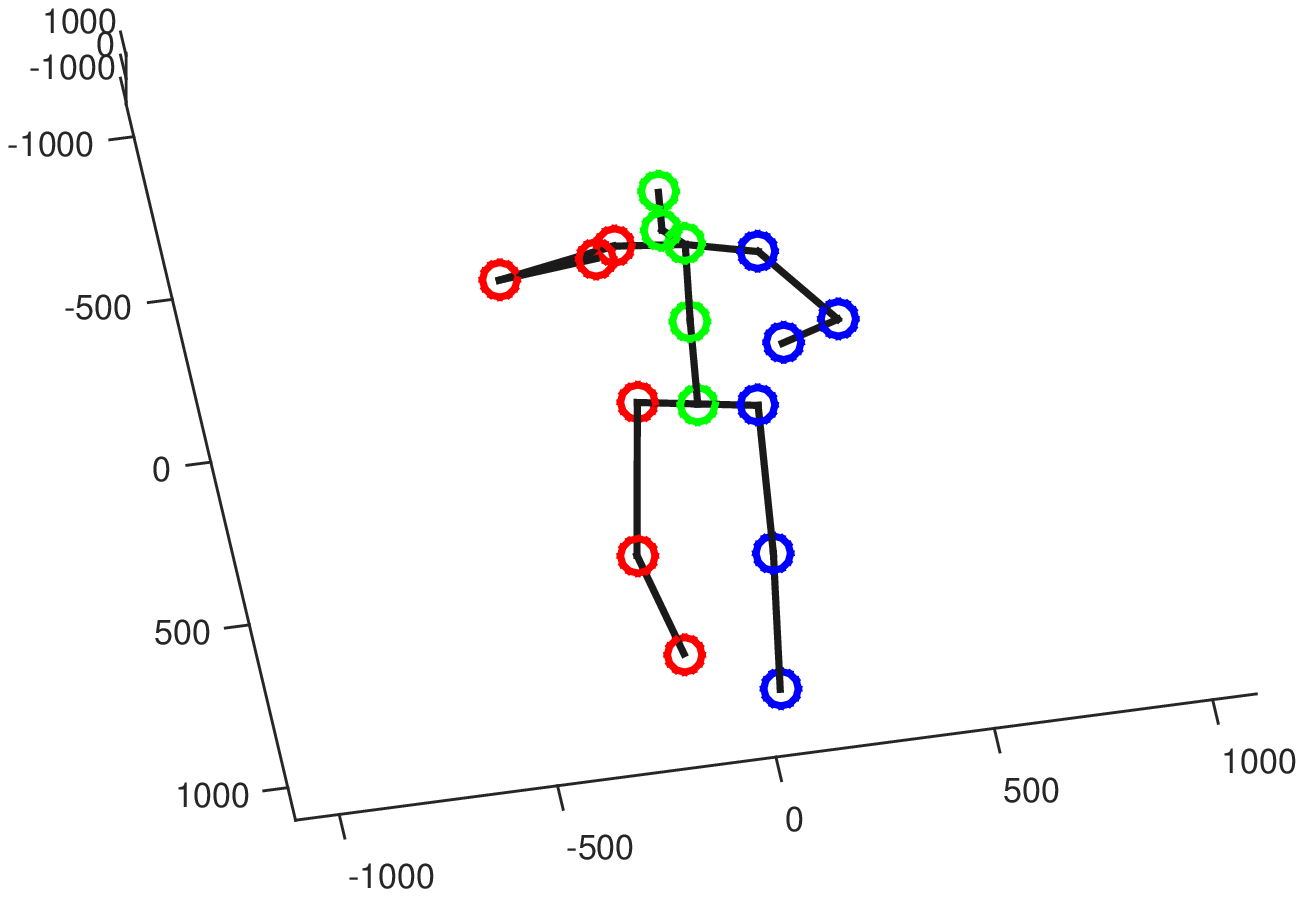}
    \end{minipage}
    \begin{minipage}{0.25\textwidth}
        \centering
        \includegraphics[width=0.99\textwidth]{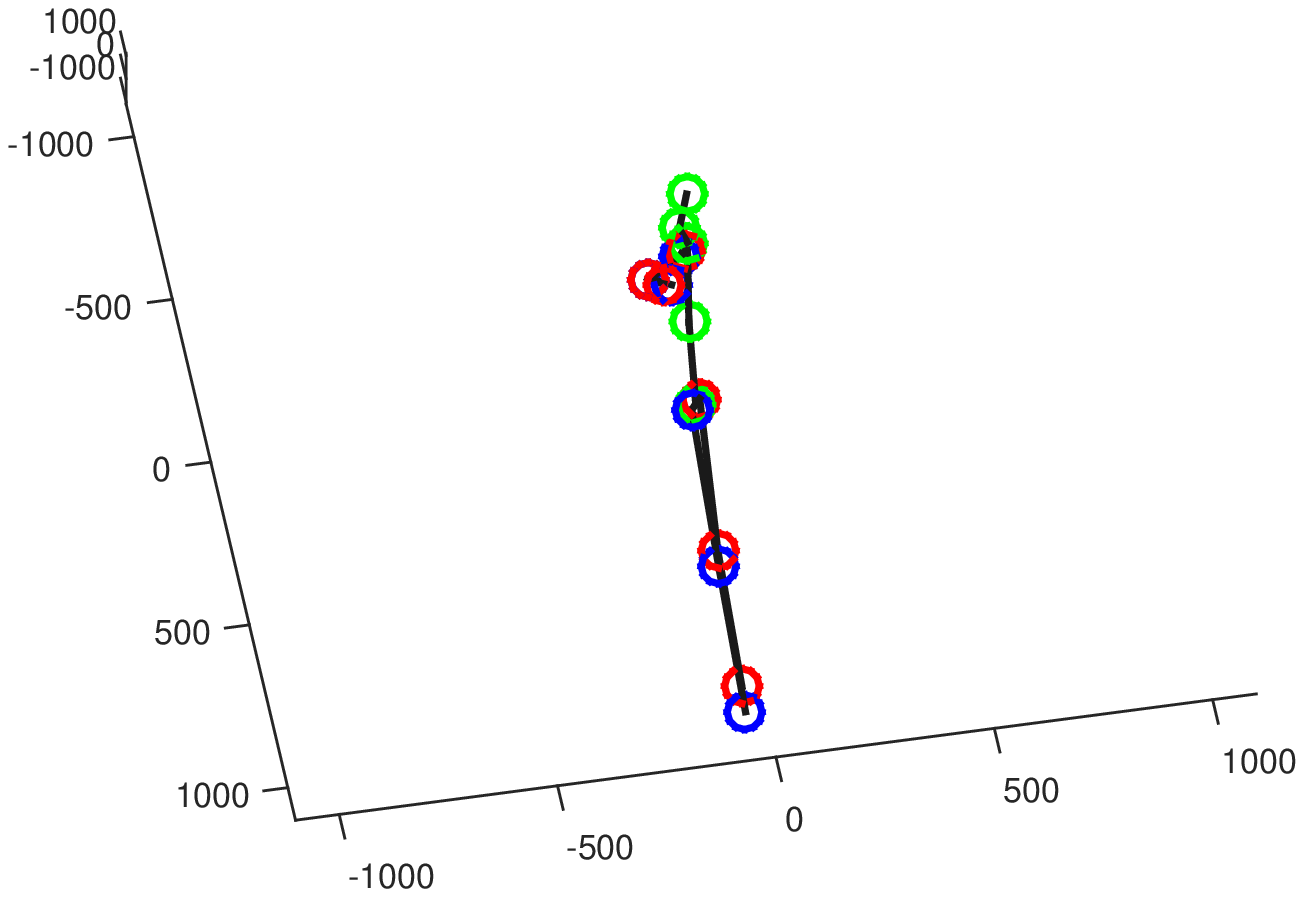}
    \end{minipage}
    \begin{minipage}{0.25\textwidth}
        \centering
        \includegraphics[width=0.99\textwidth]{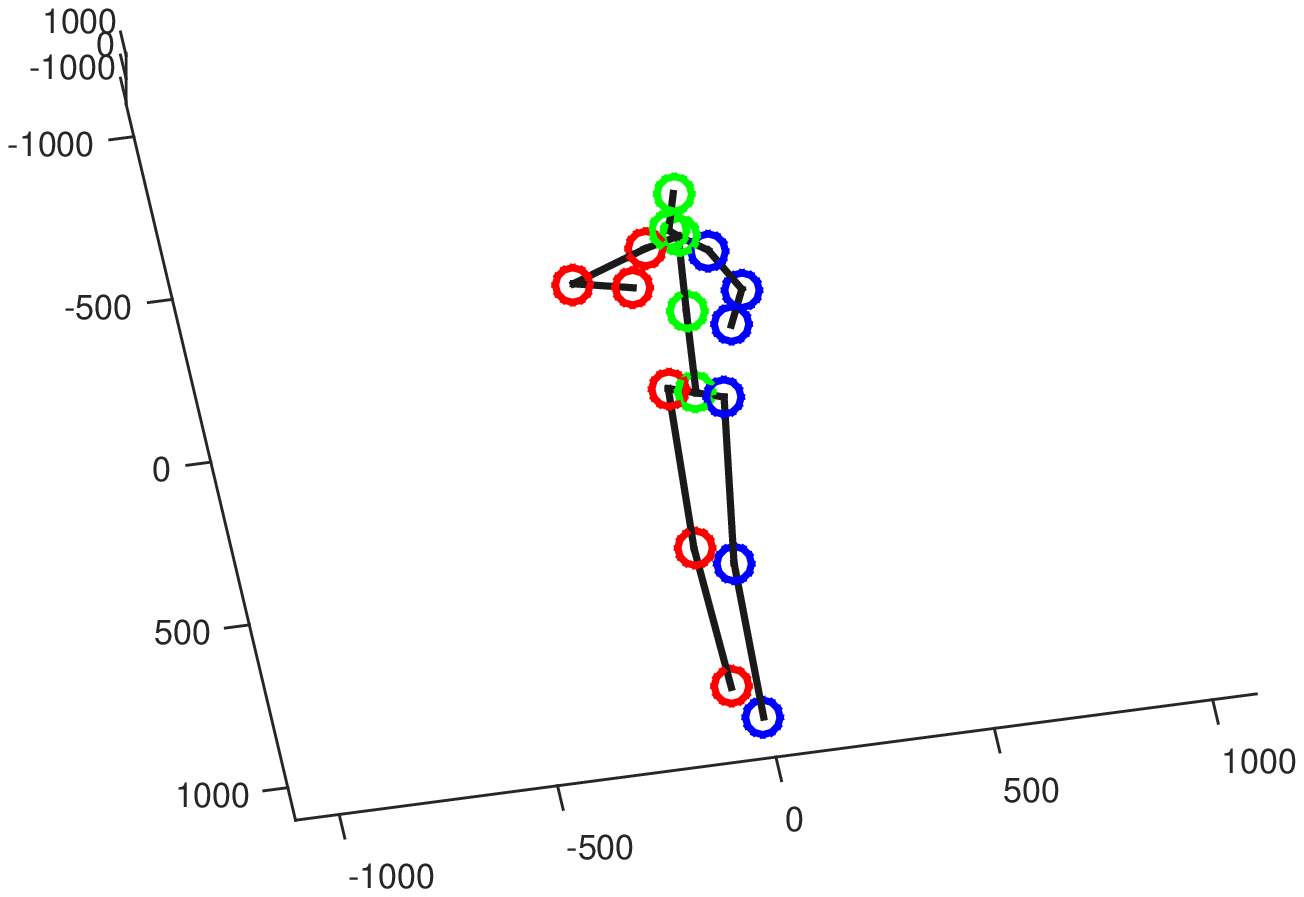}
    \end{minipage}

    \begin{minipage}{.22\textwidth}
        \centering
        \includegraphics[width=0.99\textwidth]{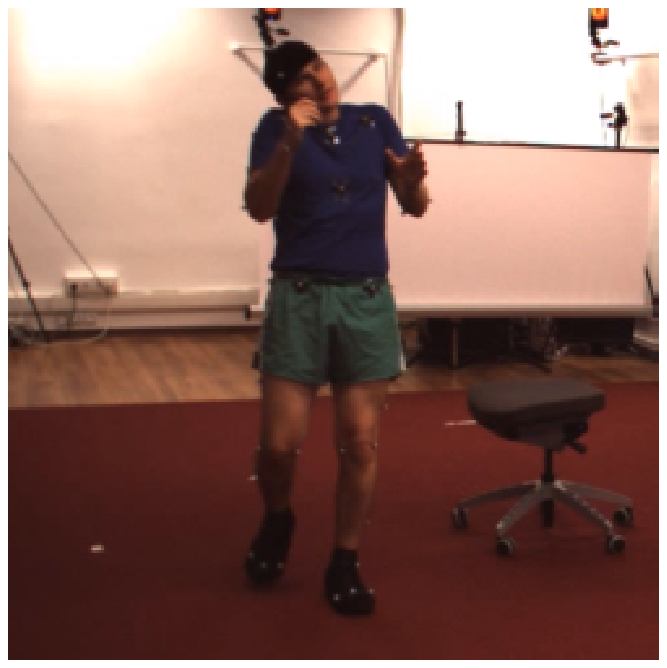}
    \end{minipage}%
    \begin{minipage}{0.25\textwidth}
        \centering
        \includegraphics[width=0.99\textwidth]{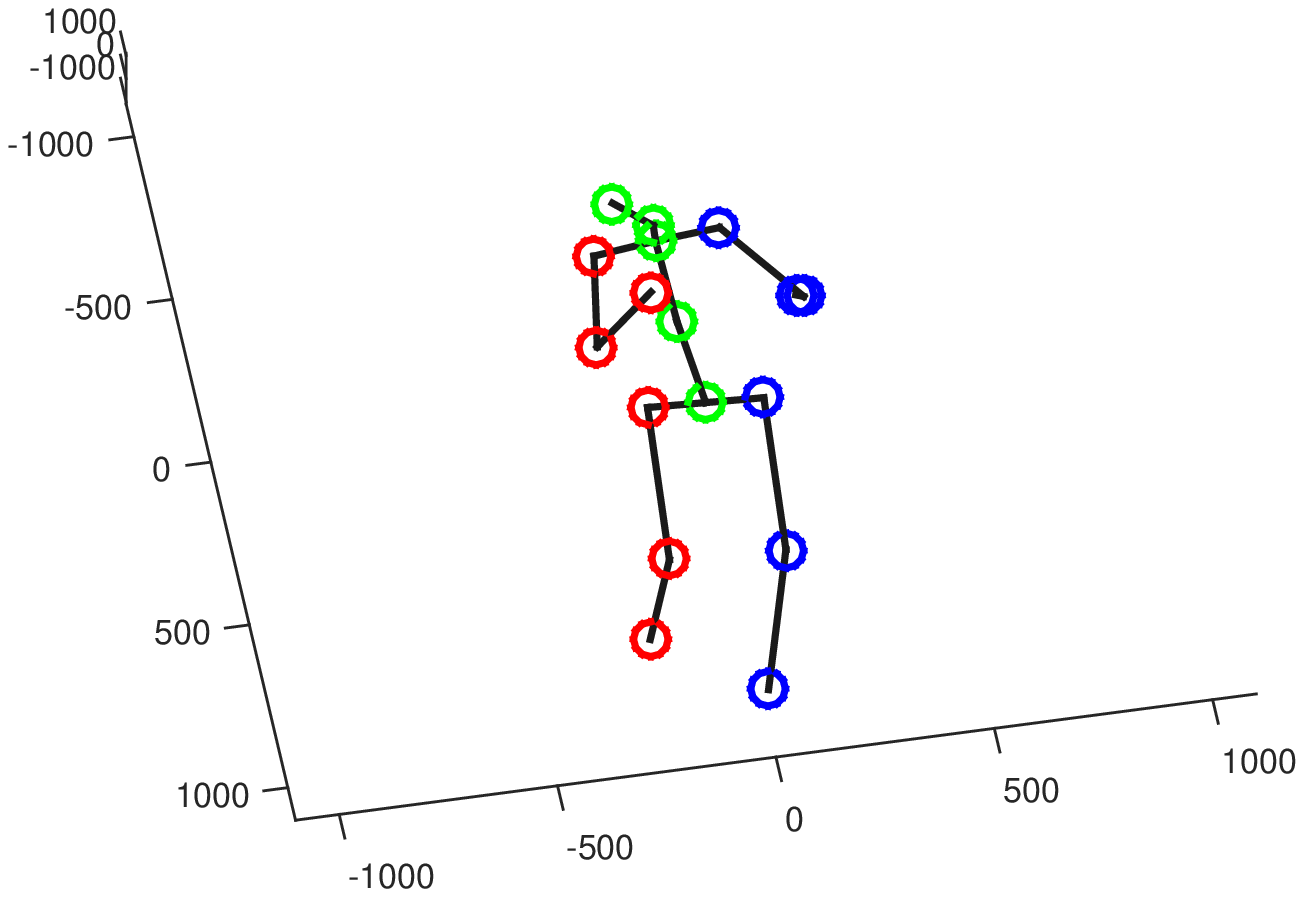}
    \end{minipage}
    \begin{minipage}{0.25\textwidth}
        \centering
        \includegraphics[width=0.99\textwidth]{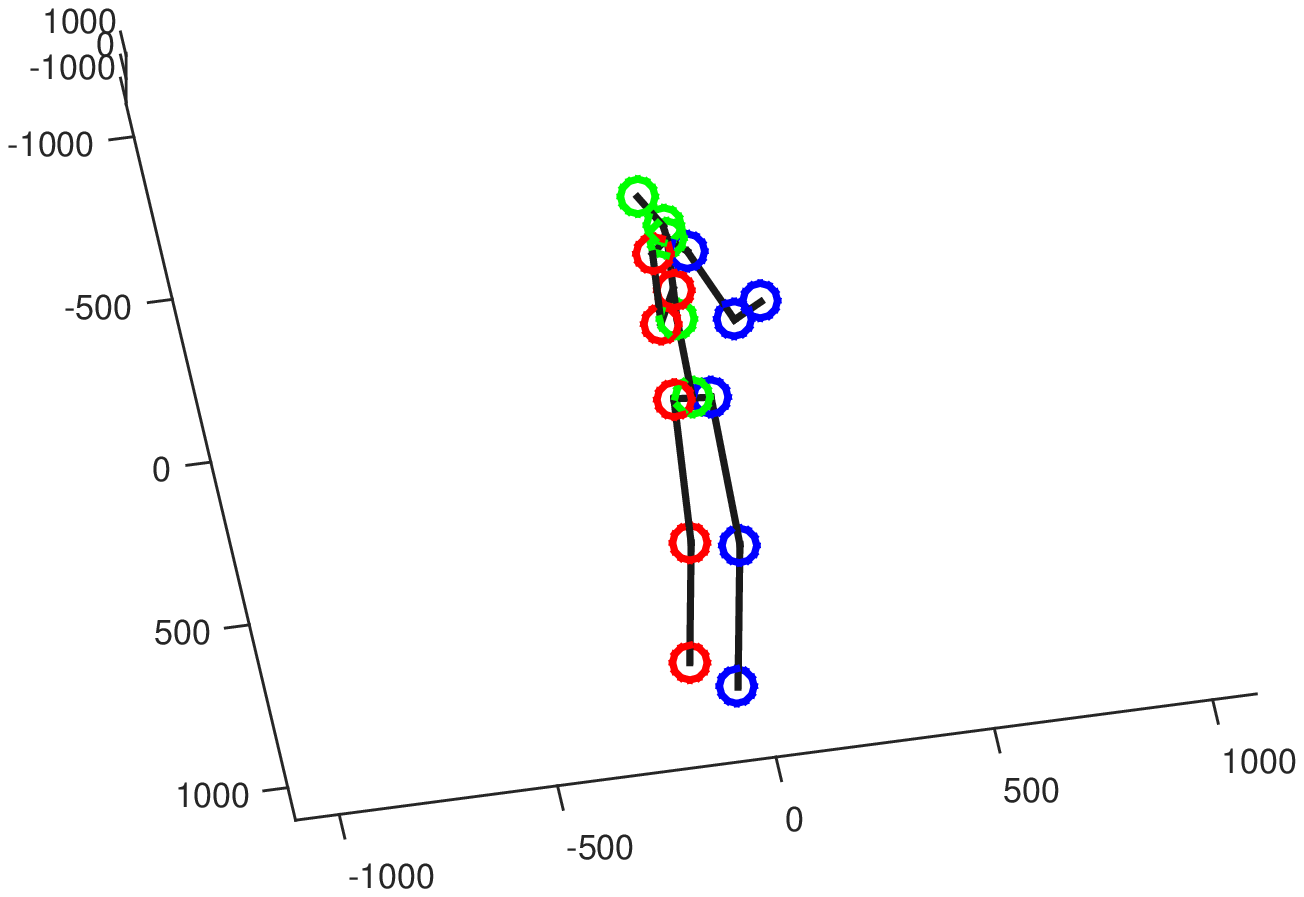}
    \end{minipage}
    \begin{minipage}{0.25\textwidth}
        \centering
        \includegraphics[width=0.99\textwidth]{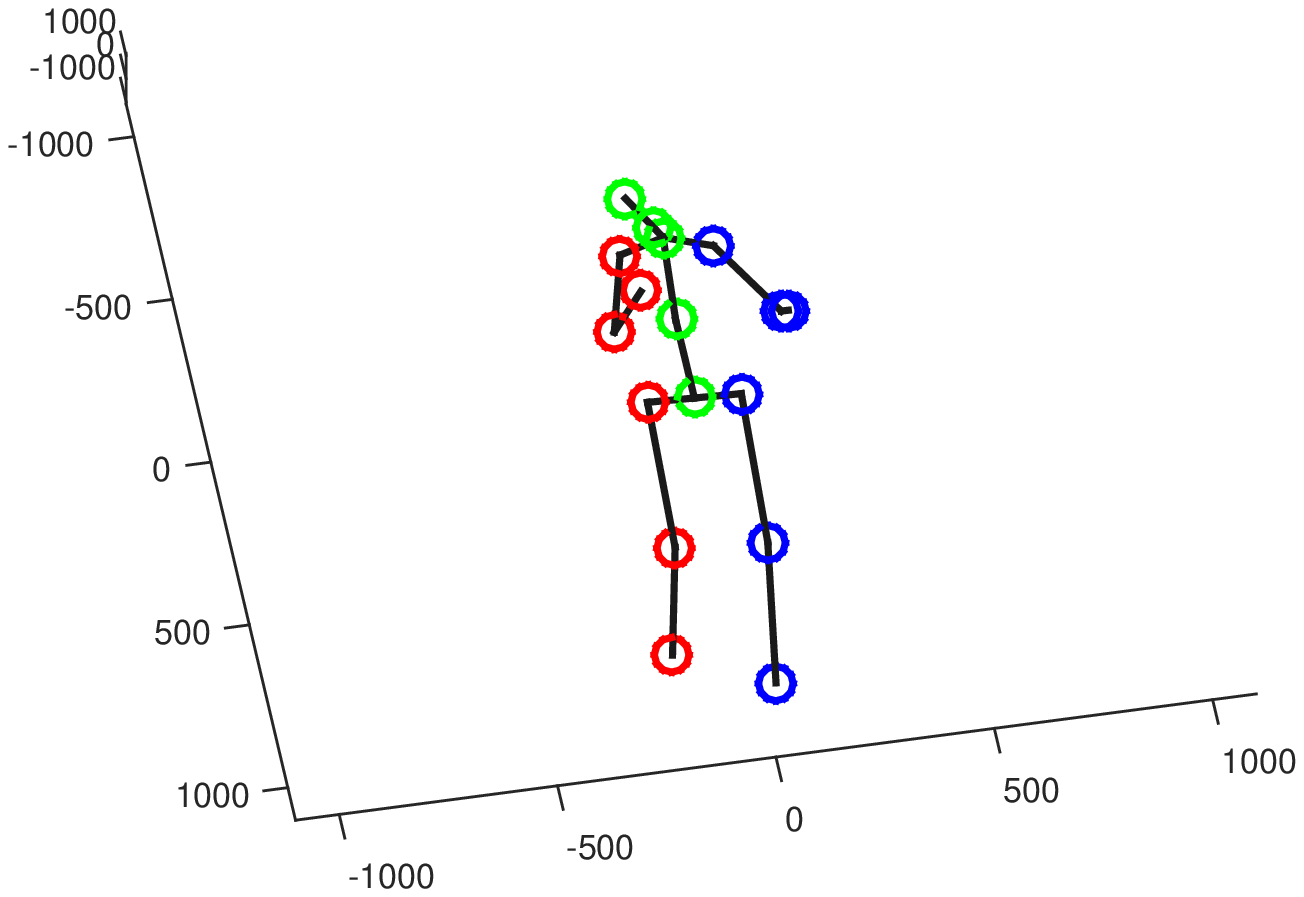}
    \end{minipage}

    \begin{minipage}{.22\textwidth}
        \centering
        Input Image
    \end{minipage}%
    \begin{minipage}{0.25\textwidth}
        \centering
        Ground Truth
    \end{minipage}
    \begin{minipage}{0.25\textwidth}
        \centering
        Without 2D info
    \end{minipage}
    \begin{minipage}{0.25\textwidth}
        \centering
        With 2D info
    \end{minipage}
  \label{fig5}
  \caption{Qualitative results of our method on Human 3.6m test dataset. The estimation results are compared with the results from the baseline method. First column : input images. Second column : Ground truth 3D position. Third column : Pose estimation result without 2D classification information integration. Fourth column : Pose estimation result with 2D classification information integration.}
\end{figure}

Finally, we illustrated qualitative results of our method in Figure~\ref{fig5}. Input images, ground truth poses, and the estimation results with and without 2D classification information are visualized. Different colors are used to distinguish the left and right sides of human bodies. It can be found that 2D pose estimation results help reducing the error of 3D pose estimation. While the CNN which does not use 2D classification information gives poor results, the estimated results are much more improved when 2D classification information is used for 3D pose estimation.

\section{Conclusions}
\label{sec:con}

In this paper, we propose novel strategies which improve the performance of the CNN that estimates 3D human pose. By reusing 2D joint classification result, the relationship between 2D pose and 3D pose is implicitly learned during the training phase. Moreover, multiple regression results with different root nodes gives an effect of ensemble learning. When both strategies are combined, 3D pose estimation results are significantly improved and showed comparable performance to the state-of-the-art methods without exploiting any temporal information of video sequences.

We expect that the performance can be further improved by incorporating temporal information to the CNN by applying the concepts of recurrent neural network or 3D convolution~\cite{tran2015learning}. Also, efficient aligning method for multiple regression results may boost the accuracy of pose estimation.

\bibliographystyle{splncs}
\bibliography{egbib}
\end{document}